\documentclass[twoside, 11pt]{article}

\usepackage{jmlr2e}

\usepackage[utf8]{inputenc}
\usepackage[flushleft]{threeparttable}
\usepackage{graphicx}
\usepackage{multirow}
\usepackage{subcaption}
\usepackage[export]{adjustbox}
\usepackage{nicematrix}

\usepackage{parskip}
\usepackage{amsmath, amssymb}
\usepackage{thmtools, thm-restate}
\usepackage{bbm}
\usepackage{booktabs}

\usepackage{hyperref}
\usepackage{enumitem}

\usepackage{appendix}
\usepackage{wrapfig}

\usepackage{lastpage}
\jmlrheading{26}{2025}{1-\pageref{LastPage}}{11/23; Revised 5/25}{7/25}{23-1590}{Xin Xu, Eibe Frank, Geoffrey Holmes}
\ShortHeadings{Multiple Instance Verification}{Xu, Frank and Holmes}
\firstpageno{1}

\newtheorem{assumption}{Assumption}

\begin{document}

\title{Multiple Instance Verification}

\author{\name Xin Xu \email xinxu75@gmail.com\\
\name Eibe Frank \email eibe@waikato.ac.nz\\
\name Geoffrey Holmes \email geoff@waikato.ac.nz\\
Department of Computer Science\\University of Waikato\\Hamilton, New Zealand
}

\editor{Tong Zhang}

\maketitle

\begin{abstract}
We explore multiple instance verification, a problem setting in which a query instance is verified against a bag of target instances with heterogeneous, unknown relevancy. We show that naive adaptations of attention-based multiple instance learning (MIL) methods and standard verification methods like Siamese neural networks are unsuitable for this setting: directly combining state-of-the-art (SOTA) MIL methods and Siamese networks is shown to be no better, and sometimes significantly worse, than a simple baseline model. Postulating that this may be caused by the failure of the representation of the target bag to incorporate the query instance, we introduce a new pooling approach named ``cross-attention pooling'' (CAP). Under the CAP framework, we propose two novel attention functions to address the challenge of distinguishing between highly similar instances in a target bag. Through empirical studies on three different verification tasks, we demonstrate that CAP outperforms adaptations of SOTA MIL methods and the baseline by substantial margins, in terms of both classification accuracy and the ability to detect key instances. The superior ability to identify key instances is attributed to the new attention functions by ablation studies. We share our code at \url{https://github.com/xxweka/MIV}.
\end{abstract}


\section{Introduction}

\begin{figure}[t]
     \centering
     \begin{subfigure}[c]{0.9\textwidth}
         \includegraphics[width=\textwidth]{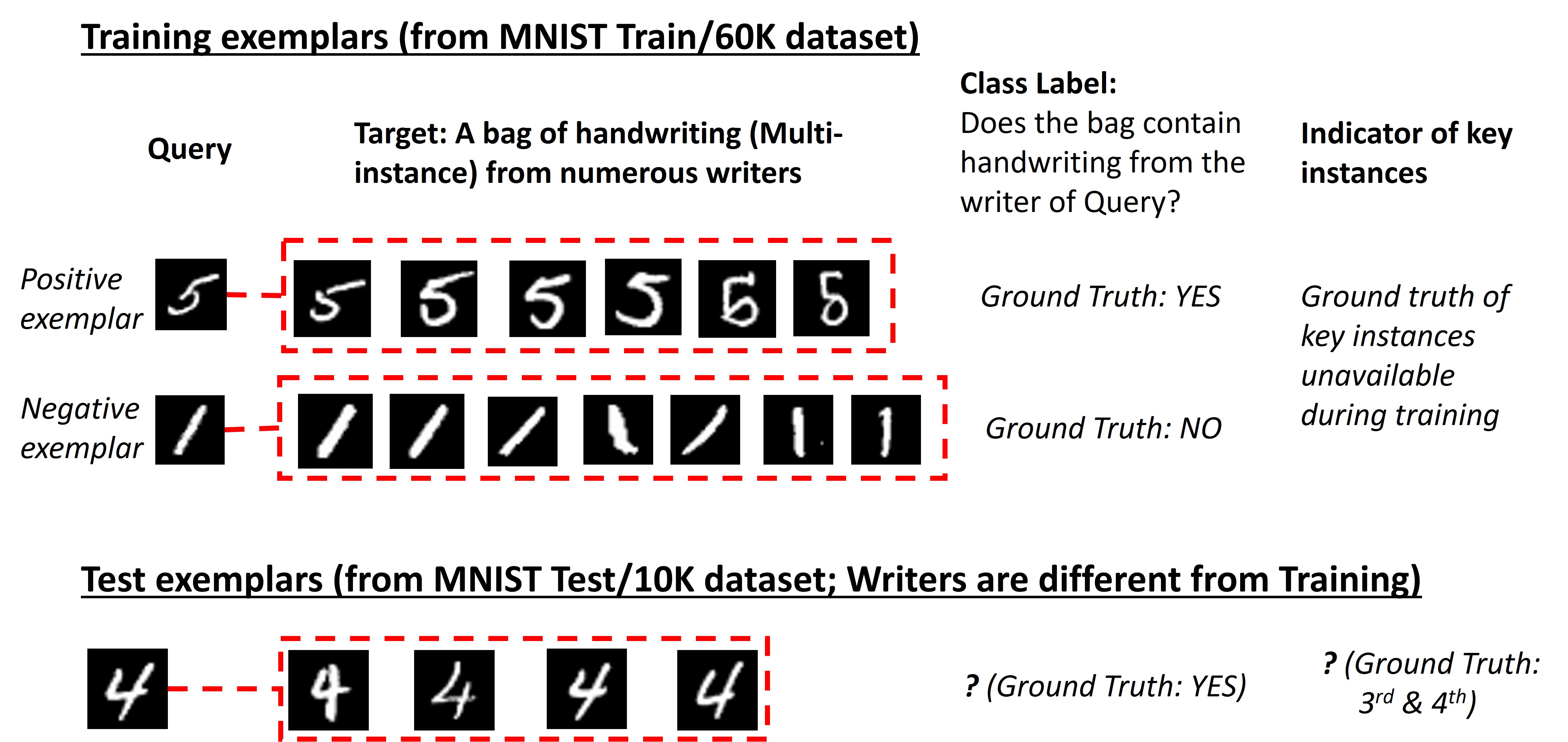} 
         \caption{} \label{task_a}
     \end{subfigure}
     \begin{subfigure}[c]{0.4\textwidth} 
         \includegraphics[width=\textwidth]{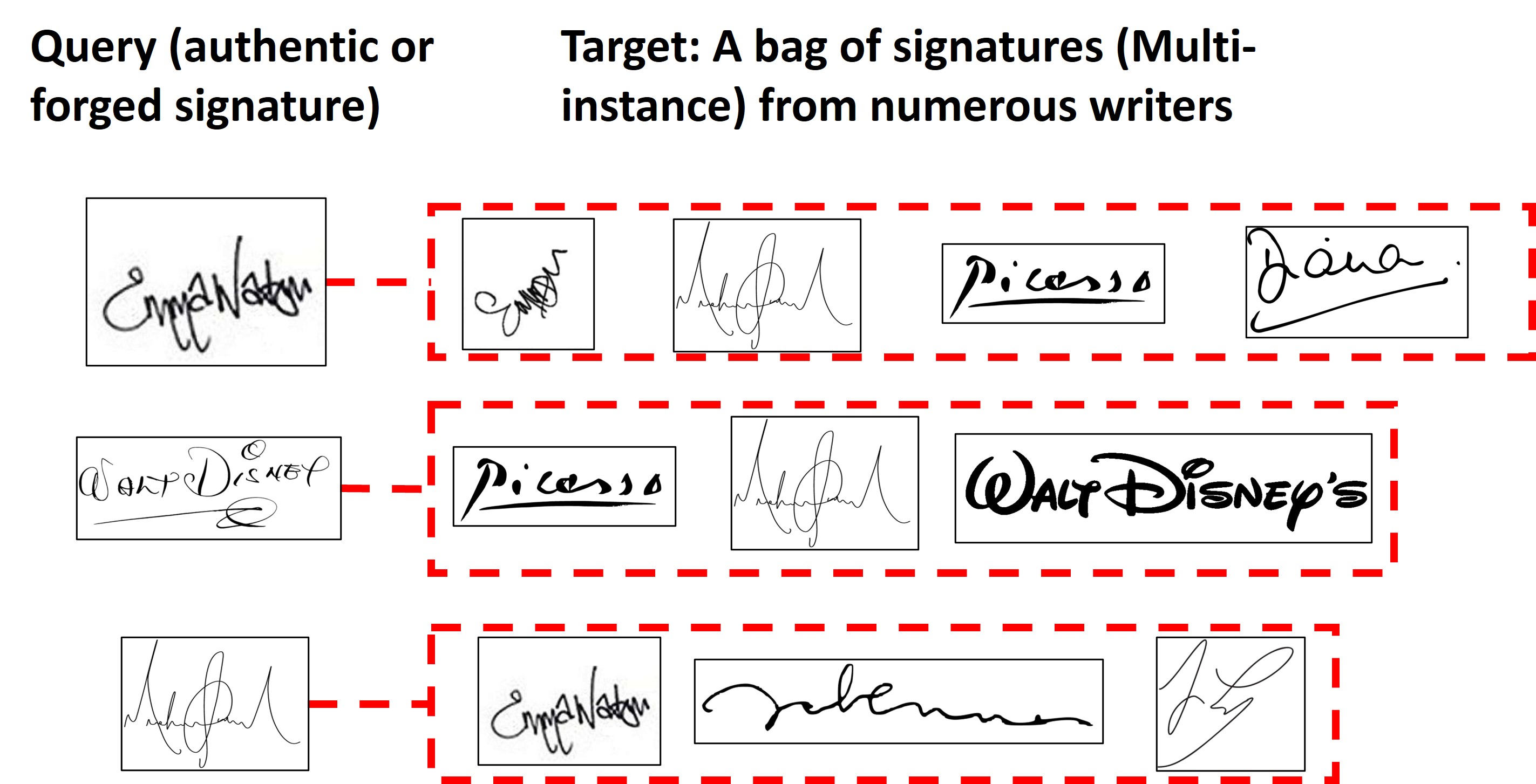} 
         \caption{}\label{task_b}
     \end{subfigure}
     \hfill
     \begin{subfigure}[c]{0.5\textwidth} 
         \includegraphics[width=\textwidth]{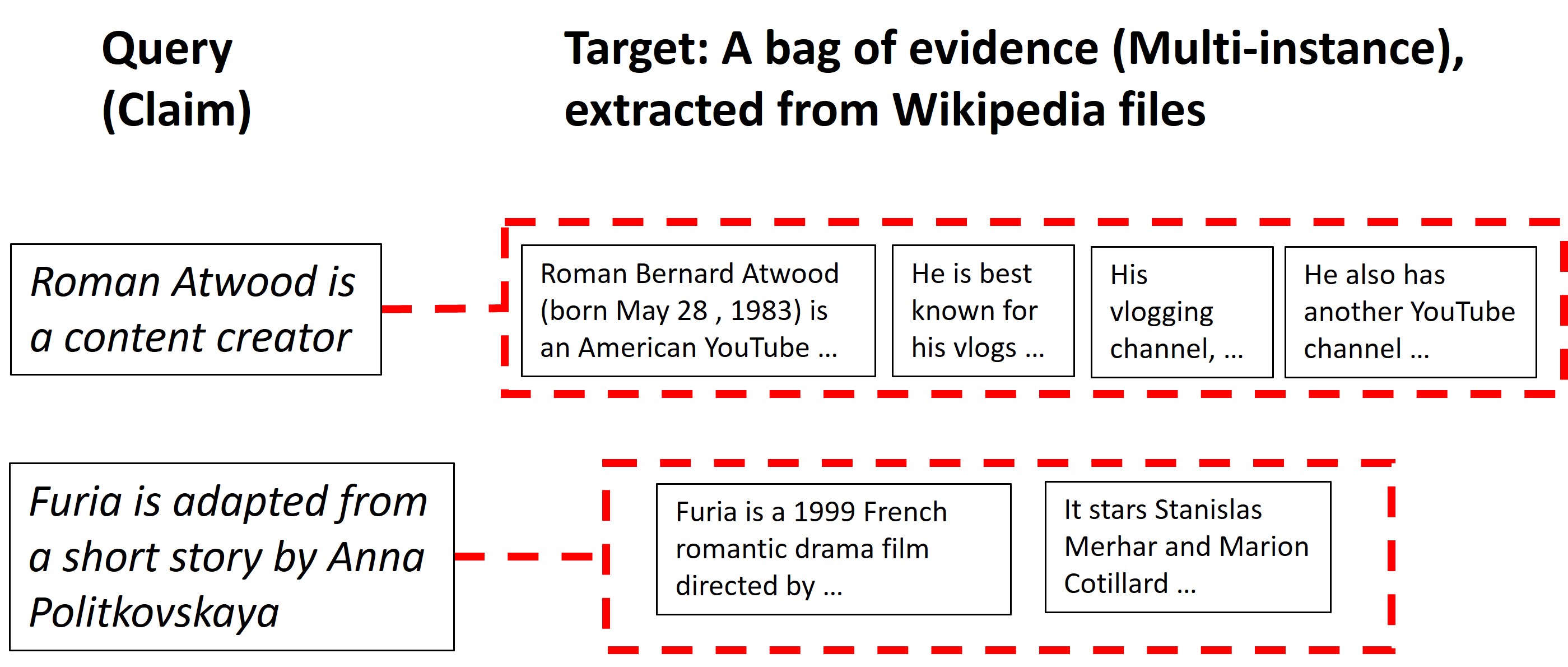} 
         \caption{} \label{task_c}
     \end{subfigure}
     \caption{MI verification tasks, with examples from computer vision (CV) and natural language processing (NLP): (\ref{task_a}) verifying handwritten digits (QMNIST); (\ref{task_b}) signature verification (displayed signatures are for illustration only, \emph{not} the actual data used in our experiments\protect\footnotemark); (\ref{task_c}) fact extraction and verification (FEVER).}
     \label{fig:miv_tasks}
\end{figure}

In multiple instance (MI) verification, each exemplar for machine learning consists of a pair of two objects---a query instance and a bag of target instances---and a binary class label that indicates whether or not the bag contains an instance pertaining to the same (unobservable) class as the query instance.\footnote{For brevity, we will refer to these components as ``query'', ``target bag'', ``exemplar label'', and ``query class'', respectively.} Figure \ref{task_a} shows an example: a synthetic task based on an extended version of the MNIST data consisting of digits and their writers' ID (\cite{qmnist19}). In this task, each exemplar contains multiple instances of the same digit written by different writers. The question is whether the target bag contains handwritten digits from the writer who wrote the query---and if there are any, which ones are ``key instances''.\footnote{We use the terms ``key instance'' and ``witness instance'' interchangeably throughout this paper, following standard practice in the literature.} Other practical applications are illustrated in Figures \ref{task_b} (signature verification) and \ref{task_c} (fact verification) respectively. Such tasks are of practical importance because they represent settings that require ``verification with noise''. While classic verification tasks involve comparing a query instance to a single target instance, real-world targets available for verification are often ``noisy'' in the sense that they consist of multiple candidates, some irrelevant, to be compared. Importantly, which individual target instances are relevant is unknown, possibly because obtaining labels indicating relevance is too costly. Hence, verification needs to be conducted using a bag of noisy target objects. Note that this generally involves establishing candidate instance relevance to the outcome of the verification process as well---a potentially very useful byproduct.

\footnotetext{Signatures in Figure \ref{task_b} are made up and for illustration only, due to data use compliance requirements.}

Conceptually, the MI verification (MIV) problem may be characterized as a combination of two sub-problems: verification (\cite{siamver93}) and multiple instance learning (MIL \cite{mil97}). Verification is frequently formulated as a binary classification task based on a pair of inputs, predicting whether or not the paired inputs are of the same class. The standard solution is to apply Siamese neural networks, as in (\cite{siamver93, siamsim05, siamese15}), but they do not have a built-in mechanism to aggregate or select multiple elements within an input, which makes them unsuitable for MIV. On the other hand, algorithms for MIL, which classify bags of instances, address the ambiguity in how a bag-level class label is related to the unknown labels of the instances in a bag (\cite{milreview17, milsurvey18}). However, existing MIL methods do not provide a dedicated mechanism to take into account the information provided by a query instance. In this paper, we argue that new algorithms and models are required for the MIV problem setting, and provide theoretical and empirical evidence showing the benefit of explicitly incorporating both components---the target bag and the query instance---and modeling their relationship during learning and inference. We revisit the distinction between MIV, standard verification tasks, and MIL in Section~\ref{sec:miv_concept}.

We also stress that ``multiple instance verification'' and similar terms used in domains such as biometrics (\cite{micnn19, mibiomgan22}) refer to problems that differ from the setting considered in our paper. In these problems, all instances of a bag, by definition, have \emph{homogeneous} unobservable labels, implying that there is no noise or irrelevant elements in the bag for verification. This is in contrast to the verification problem with \emph{heterogeneous} instances studied in this paper. Crucially, just as in traditional MIL, an exemplar label only provides information about the existence of key instances in a bag but does not identify them. Moreover, since the query instance may belong to a class that has not been observed when the verification model was trained, it enables classification into previously unknown classes. Following the rationale of \cite{siamese15}, this is analogous to an MI version of zero-shot classification.

A naive approach to MIV is to combine existing verification and MIL methods as building blocks, for example, by creating a bag-level representation for the target bag using a state-of-the-art (SOTA) MI pooling method, and then treating this bag-level representation together with the query instance as the two inputs to a Siamese network for verification. Suitable SOTA MI pooling methods include approaches based on gated attention (``Gated-Attention-Based MIL'' \cite{gabmil18}), the transformer decoder (``Pooling by Multi-head Attention'' \cite{pma19}), and the transformer encoder (e.g.,~\cite{transmil21}). Alternatively, non-attention-based methods such as MI-Net (\cite{minet18}) and bi-directional LSTM (Bi-LSTM \cite{mil-lstm20}) can be applied. However, we show that solutions based on this straightforward approach, regardless of which of the above MIL algorithms is applied, yield unsatisfactory classification performance. More specifically, a simplistic ``baseline'' model is shown to outperform the resulting instantiations---henceforth referred to as ``benchmarks''---of the naive approach to MIV.

We postulate that the benchmarks perform unsatisfactorily because they do not incorporate the query instance into the bag-level representation of the target bag: the ``query'' concept does not exist in the underlying MIL methods. Indeed, we show theoretically, using a probabilistic interpretation of attention scores, that failure to incorporate the query instance induces less effective attention scores. Hence, inspired by the ``query-key-value'' attention mechanism of the transformer decoder by \cite{transformer17}, we propose a new pooling approach called ``cross attention pooling'' (CAP) that creates the bag-level representation of a target bag through a cross-reference attending to the query. This means that the bag's representation is dynamic, because it varies when a different query instance is present, even if the bag itself does not change. In contrast, standard MIL methods create a static bag-level representation.  CAP also enables the introduction of two novel attention functions, namely ``distance-based attention'' and ``variance-excited multiplicative attention'', respectively, to address the challenge that the instances in a bag may be difficult to distinguish. Both functions accommodate multi-head formulations popular in transformers.

Despite being inspired by ``transformer-style'' attention, there are significant novelties in CAP's cross-attention mechanism that cannot be achieved by the standard attention mechanism of transformers. In particular, CAP's cross-attention facilitates two outputs, by sharing parameters between the two processes producing the outputs alongside attention scores. It also allows for a ``co-excitation'' mechanism that yields consistent improvements in performance (\emph{cf.} Section~\ref{sec:ablation}). Finally, CAP incorporates new attention functions accounting for in-bag, channel-wise variations, not seen in standard transformers' attention. All these designs make CAP's cross-attention a bespoke mechanism tailored to MIV, and well-suited for MI pooling. Importantly, simply adapting the commonly used ``transformer-like'' cross attention---a mechanism intended to construct contextualized embeddings---to the MIV setting (\emph{cf.} Section~\ref{sec:pma}) may fail to effectively tackle this problem, even producing unintended results, see Section~\ref{sec:experiments}.

In addition to achieving high classification accuracy, the ability to identify key instances in a target bag is also highly desirable for MIV. This ability is sometimes referred to as ``explainability'' or ``interpretability'' in the literature on MIL, including (\cite{gabmil18, transmil21, milinterpret22}). Following recent related work (\cite{explain21, wsol23}), we quantitatively evaluate key instance detection for the methods in our study. This is possible for datasets with ground-truth identifiers of key instances.

Our experimental results, obtained from three different tasks and considering both classification and key instance detection, show that CAP is an effective solution to the MI verification problem. In terms of (bag-label) classification performance, it significantly outperforms the benchmarks and the baseline by large margins. Furthermore, using quantitative measures of key instance detection performance, we show that it identifies key instances much more accurately than both the benchmarks and the baseline. Higher classification and key instance detection accuracy are observed with different numbers of training exemplars and with varying bag sizes in the training data. We also present an ablation study showing that the components of the novel attention mechanisms responsible for delivering superior quality of key instance detection---and thus leading to better bag-level representations---are the source of better classification performance.

Our main contributions are as follows. First, we formally state the multiple instance verification problem, and create a new solution, CAP, that outperforms SOTA MIL methods combined with Siamese networks and a simple baseline method on three tasks from the computer vision (CV) and natural language processing (NLP) domains. Furthermore, we propose two new attention functions for CAP: (a) distance-based attention, which differs from the mainstream functional forms of attention because it is neither additive nor multiplicative, and (b) variance-excited multiplicative attention, which is a new form of multiplicative attention. We show that they are better able to identify key instances in MIV, resulting in better bag-level representations. While attention-based methods are not uncommon in MIL-related domains, research studying different types of attention mechanisms and analyzing their efficacy under specific conditions is rare. Within this context, we provide insights that highlight the importance of quantitative evaluation of the ability to detect key instances, which helps identify spurious model performance. We find that MI models may identify key instances incorrectly but still classify the corresponding bag correctly, and vice versa. Our work also contributes to the specific application areas we consider. For example,  in the natural language inference (NLI) subtask of FEVER, the existing approach is a two-step pipeline: evidence retrieval and support verification. Our approach combines them into a single step through CAP, and thus enables end-to-end training of the whole pipeline.

\section{Related work} \label{sec:lit_rev}

The most closely related existing work to multi-instance verification (MIV) is research on classic verification using Siamese neural networks and the deep MIL literature. The relationships between MIV tasks and problems tackled using learning approaches from these two areas are discussed further in Section~\ref{sec:miv_concept}.

The verification problem, and its solution using Siamese networks, was first introduced by the pioneering work of \cite{siamver93}. \cite{siamsim05} further explored Siamese networks for learning similarity metrics. Since then, verification problem settings have been generalized in both CV (e.g., \cite{siamcnn15}) and NLP (e.g., \cite{infersent17}) domains where Siamese networks can be applied. There are also several areas in machine learning that apply Siamese networks to solve problems similar to verification, including one-shot learning (\cite{siamese15, sce19}), self-supervised contrastive learning (\cite{simclr20, barlow21}), embedding pre-training (\cite{infersent17, sbert19}), to name a few examples.

We note that there exist approaches other than Siamese networks that may also be applied to MIV in specific application scenarios, such as using concatenation of the query instance and the target bag in an NLP task considered in \cite{fever_mil21}. We do not consider this approach here because it needs to be used jointly with other NLP techniques and cannot be trained end-to-end. Within the scope of this paper, we only consider Siamese networks, which provide a more generic and domain-agnostic approach.

Regarding MIL, whose study began with the seminal work of \cite{mil97} and which can be viewed as a form of ``weakly supervised learning'' (see~\cite{milreview17}), there are strong connections to our work. Loosely speaking, depending on whether a method (1) explicitly creates a bag-level representation or (2) directly aggregates instance-level scores (or logits), MIL methods may fall into two categories: bag-level or instance-level approaches (see \cite{milsurvey18}). While many methods from earlier MIL research are instance-level (e.g., \cite{emdd02, miboost04}), later research typically focused on bag-level methods (e.g., \cite{milsoftbag15, minet18, mil-lstm20}). More recently, mainstream deep MIL research has been dominated by attention-based methods (including \cite{gabmil18, pma19, transmil21}), which can be viewed as both bag-level and instance-level methods. The attention scores can be viewed as a weighting function that not only aggregates instance-level representations into bag-level ones, but aggregates instance-level scores as well. These methods are the focus of our study and are used to derive the benchmark models we use.

There are also sub-domains of MIL that indirectly relate to our paper. One strand of research considers dynamic representations of bags based on modeling the context of instances (\cite{milcorr09, milsoftbag15, mildp18}). Unlike our study, this research focuses on modeling contextual information \emph{within} a bag, rather than information from \emph{outside}. Another strand emerging more recently considers the quality of key instance detection in weakly supervised learning, proposing evaluation metrics that do (\cite{explain21, wsol23}) or do not (\cite{milinterpret22}) require unobservable instance-level labels. We focus on the former metrics in our study and use them to measure the attention scores' ability to identify key instances. Our goal is to quantitatively evaluate these scores and draw insights in the context of our theoretical propositions in Section 3.4. 

Like recent MIL (\cite{pma19, transmil21}) and other research (\cite{vit21, sbert19}), our methods are also strongly inspired by the transformer architecture (\cite{transformer17}) and can be grouped with the wider range of research that is based on this seminal model. Indeed, many CV or NLP problems solved by transformers can be represented as MIL problems with sequentially ordered instances, in which the sequence order can be addressed by the addition of a positional embedding (see, for example, \cite{transmil21}).

Finally, there are specific areas of CV and NLP that are related to our work. Examples in CV include weakly-supervised object detection or localization (\cite{explain21, wsol23}) and one-shot object detection (\cite{sce19}). In NLP, one area is ``fact extraction and verification'' (FEVER,~\cite{fever18}). It is well-suited to the MIV framework because of the framework's ability to deal with ambiguity. Indeed, we evaluate our approach on one of the FEVER datasets.

\section{Multi-instance verification (MIV): problem definition and an interpretation of attention scores} \label{sec:theory}

The scenario of ``verification with noise'' (or ``verification with unobservable instance labels'') is applicable to many tasks in the areas of CV and NLP. To give a few examples, practical use cases of MIV may include:
\begin{itemize}
\item verification of whether documents (bags of unlabeled paragraphs) contain content generated by a particular writer based on the query representing a writing sample of this writer;
\item verification of whether bags of unlabeled images contain targets such as people, endangered species, plants, etc., specified by a reference image (query). This use case can be generalized to bags of sequentially ordered images, like video clips; if the sequence order is essential, it can be addressed by the addition of positional embeddings;
\item verification of whether histology whole slide images (bags of unlabeled patches) indicate certain diseases based on a query image of diagnosed (tumor) cells. A generalization of this use case relates to the broader subject of ``weakly-supervised one-shot object detection'' in the CV domain;
\item verification of whether network logs (bags of unlabeled scripts or time series) record traces of cyber-attacks based on the footprint of a past attack (query), and so on and so forth.
\end{itemize}

It is worth noting that standard information retrieval tasks, i.e., searching for content relating to a query, can usually be addressed by classic verification methods, \emph{without} the MIV formulation. However, when the ground truth for standard retrieval tasks is unavailable, that is, it \emph{cannot} be easily defined whether a unit of searched content ``matches'' the query, the standard training procedures may be impermissible due to lack of class labels. Under those non-standard circumstances, if a label is available for a bag of (unlabeled) units of content---i.e., a label of whether the bag contains a ``match'' while the precise match is unknown---MIV may be applicable. For example, the first two examples above---document retrieval based on the writing style of a sample, or searching videos based on a single image of an object---are relevant use cases. Hence, in terms of practical applications, MIV and standard retrieval problems apply to \emph{mutually exclusive} scenarios, thus \emph{complementing} each other. The appropriate formulation should be adopted according to available data.

\subsection{Three example MIV tasks} \label{sec:tasks}

To better understand MIV and its usefulness, we first present detailed information on the three MIV tasks illustrated in Figure \ref{fig:miv_tasks}. These are tasks for which we were able to obtain publicly available data for our experiments.

To generate the exemplar labels for handwriting verification (Figure \ref{task_a}), we used an extended version of MNIST, the QMNIST dataset (\cite{qmnist19}), which includes the writer IDs of all the handwritten digits in the MNIST dataset. To differentiate our task from the standard MNIST classification task of digit recognition, and to focus on the query class (i.e., the writer ID), we constructed each exemplar using the same digit across the entire exemplar comprising the query and the target bag. To avoid trivial cases where the query is also in the bag, we ensured that all images in the bag are different from the query even if they may be written by the same writer. Note that the ``inference'' stage in Figure \ref{task_a} depicts the two goals of our study: verification and key instance detection. 

In the signature verification task (Figure \ref{task_b}), a target bag is composed of multiple authentic signatures, usually called ``anchor'' images. There are three types of query signatures: 
\begin{enumerate}[leftmargin=*]
    \item an authentic signature that can be verified by some anchors in the target bag (i.e., the query and anchors are written by the same writer),
    \item a professionally forged signature that looks similar to some anchors, and
    \item an unmatched signature that does not look close to any anchor.
\end{enumerate}
The (unobservable) query class is the writer ID. The exemplar's label signals whether or not the query can be verified as authentic by any anchor in the target bag. Thus, it is ``1'' for the first case and ``0'' for the second and third cases considered in the above list.

In the FEVER task (Figure \ref{task_c}), the target bag consists of multiple pieces of evidence, each represented by paragraphs extracted from various Wikipedia pages, and the query is a claim about a fact. The (unobservable) query class is the semantic category of the query paragraph (determined by human experts): supportive of the claim or unsupportive. By construction, any query is ``supportive'' of itself. If a piece of evidence in the target bag is supportive, the query is said to be supported by the evidence, i.e., they are of the same query class. Thus, the exemplar's label is whether or not the query is supported by any evidence in the target bag.

For all three tasks described above, we were able to collect identifiers of the key instances, either by construction or from raw data. This information is only used during testing, to evaluate models' performance in detecting key instances, but \emph{not} during training.

\subsection{Formal definition of MIV} \label{def_miv}
To formally state the task of MIV, we use the following notation. The (unobservable) class label of an instance $x$ is denoted by $y\in c$ where $c=\{1, 2,\cdots, L\}$ and $L$ is the number of unobservable classes. Note that $L$ can be $\infty$ in theory. The query instance and its unobservable class label are denoted by $x_Q, y_Q$, and the bag of target instances and their unobservable class labels by $\{x_1, x_2, \cdots, x_N\}$ and $\{y_1, y_2, \cdots, y_N\}$ respectively. Unless stated otherwise, in this section, we assume the instances, $x_Q$, $x_n(n=1,\cdots, N)$, have all been embedded in an $F$-dimensional feature space, that is, $\in\mathbb{R}^F$. 

\begin{definition}[MI verification exemplar] \label{def1}
An MI verification exemplar is a tuple, \hfill\break $(x_Q, \{x_1, x_2, \cdots, x_N\})$, with a label $Y$ given by 
\[
Y = 
\begin{cases}
    0,  & \emph{iff } y\neq y_Q\quad \forall y\in\{y_1, y_2, \cdots, y_N\}\\
    1,  & \emph{otherwise.}
\end{cases}
\]
\end{definition} 

It should be noted that an MIV task can be converted into a ``multiple pairs of instances'' learning problem, analogous to MIL, by pairing the query instance with each instance in the target bag. We state this formally as a lemma below.

\begin{restatable}[Conversion of MIV exemplars]{lemma}{fstlemma} \label{lemma1}
An MI verification exemplar in Definition \ref{def1} can be represented as a bag of instance-pairs: $\{(x_Q, x_1), (x_Q, x_2), \cdots, (x_Q, x_N)\}$, each with an unobservable binary class label $\hat y_n\in\{0,1\}, n=1,2,\cdots, N$ defined as
\[
\hat y_n = 
\begin{cases}
    0,  & \emph{iff } y_n\neq y_Q,\quad n=1,2,\cdots,N\\
    1,  & \emph{otherwise.}
\end{cases}
\]
The bag-level class label is given by
\[
Y = 
\begin{cases}
    0,  & \emph{iff } \sum_n^N \hat y_n = 0\\
    1,  & \emph{otherwise.}
\end{cases}
\]
\end{restatable}
The proof is obtained trivially by construction.

\subsection{Permutation invariance in MIV} \label{def_perminv}

In MIV problems, like traditional MIL, the order of the instances in a bag should not matter---the function used to score bags of data should be permutation-invariant. Lemma \ref{lemma1} allows us to borrow theory from existing MIL literature, e.g.,~\cite[Theorem 1]{gabmil18}, to provide a sufficient and necessary condition for the scoring function, $S(X)$, of a bag of instance pairs to be permutation invariant. We state this formally in Lemma \ref{lemma2}, simply replacing an instance $x_n$ with a pair of instances $(x_Q, x_n)$ in \cite[Theorem 1]{gabmil18}.

\begin{lemma}[Condition for permutation invariance] \label{lemma2}
A scoring function, $S(X)\in \mathbb{R}$, for a set of pairs $X=\{(x_Q, x_1), (x_Q, x_2), \cdots, (x_Q, x_N)\}$, is permutation invariant to the elements in $X$, if and only if it can be decomposed in the following form
\begin{align} \label{perm_inv}
S(X) &= g\Big(\sum_{(x_Q,x_n)\in X}f(x_Q,x_n)\Big), \qquad n=1,2,\cdots,N
\end{align} 
where $f$ and $g$ are suitable transformations.
\end{lemma}

Lemma \ref{lemma2} allows us to demonstrate that the models under the CAP framework, introduced in Section \ref{sec:cap}, are permutation invariant. As we will see, the models in Section \ref{sec:cap} use a scoring function that can be written as
\begin{equation}
    S(X) = H(x_Q) diag(\alpha)\underset{j=1,\cdots,h}{ concat}\Big(\sum_n^N \big(A_j(x_Q, x_n)G_j(x_n)\big)\Big)^T,\nonumber 
\end{equation}
where $\underset{j=1,\cdots,h}{concat}$ concatenates $h$ vectors (one from each attention head), $A_j(x_Q, x_n)\in\mathbb{R}$ is a scalar attention score for the $n^{th}$ instance in the target bag at the $j^{th}$ head, $H(x_Q)\in\mathbb{R}^{1\times F}, G_j(x_n)\in\mathbb{R}^{1\times\frac{F}{h}}$ are transformations of $x_Q$ and $x_n$ (at the $j^{th}$ head) respectively, and $diag(\alpha)\in\mathbb{R}^{F\times F}$ is a diagonal matrix of learnable parameters.

We can split the vectors $H(x_Q)$ and $\alpha$ into $h$ equal-length segments, and rewrite this scoring function by moving $H(x_Q)$ and $diag(\alpha)$ inside the summation,
\begin{align}\label{perminv_proof}
S(X) &= \sum_j^h \bigg[\sum_n^N A_j(x_Q, x_n)\Big(G_j(x_n) diag(\alpha_j)H_j(x_Q)^T\Big)\bigg] \nonumber\\
&= g\Big(\sum_n^N f(x_Q, x_n) \Big),
\end{align} 
where $H_j(x_Q)\in\mathbb{R}^{1\times\frac{F}{h}}, \alpha_j\in\mathbb{R}^{\frac{F}{h}}$ denotes the $j^{th}$ segment of the corresponding vectors. 

The first line of Equation~\eqref{perminv_proof} states that the logit of our models is the sum of ``sub-logits'' from multiple heads. The sub-logit from each head is the attention-score weighted sum of $N$ Siamese-twin similarities computed from $N$ pairs, where each pair is formed by the query (transformed by $H$) and the corresponding instance in the target bag (transformed by $G$)---conforming to the conversion process in Lemma \ref{lemma1}. The last line is obtained by simply treating $\sum_j^h$ as $g(\cdot)$, and everything inside the summation $\sum_n^N$ as $f(\cdot, \cdot)$. 

Although with theoretical merits, the results in this section are not particularly helpful in guiding us to find specific functional forms of $S(X)$. Indeed, even naive employment of existing SOTA MIL pooling mechanisms may fit  Lemma \ref{lemma1} and \ref{lemma2}. However, intuitively, this naive approach may be problematic because these mechanisms do not account for the special role of the query in attention. To substantiate this intuition, we need a new framework for interpreting attention scores.

\subsection{A probabilistic interpretation of attention scores} \label{sec:as}

We present a probabilistic interpretation of attention scores based on the observation that they are non-negative and sum to 1. Therefore, they form a probability distribution $Pr(U)$ of a discrete random variable $U\in\{n\mid n=1, 2, \cdots, N\}$ that encodes the relevance to the exemplar label of the $N$ instances it contains. This enables the use of two concepts from probability theory. First, the attention-weighted sum of the instances' feature vectors can now be seen as an expectation, where this expected vector represents the entire bag. Second, the entropy $H(U)$ of $Pr(U)$ can be used to measure the informativeness of a set of attention scores---large entropy corresponds to an even distribution of attention scores, which provides little information about the relevance of individual instances to the bag's label---and well-known properties from information theory (\cite{eit06}) can be applied.

The probability distribution for a particular MIV exemplar clearly depends on the information in the query instance and the target bag. We assume that the query instance and the instances in the target bag have all been embedded in an $F$-dimensional feature space, yielding $V^{Query}\in\mathbb{R}^{F}$ and $V^{Target}\in\mathbb{R}^{N\times F}$. 
It can be shown that, under mild conditions, we can \emph{strictly} reduce uncertainty, and thus \emph{strictly} increase informativeness regarding the key instances in a target bag, by considering information from both sources, $V^{Target}$ and $V^{Query}$, rather than just one. We state this as Proposition~\ref{prop1}, together with Assumption~\ref{ass1}, as follows.

\begin{assumption}[Uncertainty of conditioning variables' informativeness] \label{ass1}
\begin{align}
Pr\Bigg(H\Big(U\mid V^{Query}\Big) > H\Big(U\mid V^{Target}\Big)\Bigg)>0, \\ \nonumber
Pr\Bigg(H\Big(U\mid V^{Query}\Big) < H\Big(U\mid V^{Target}\Big)\Bigg)>0. \nonumber
\end{align}

\end{assumption}
Assumption \ref{ass1} states that informativeness of attention scores brought by $V^{Query}$ and $V^{Target}$ should be uncertain, that is, there should be no guarantee that one variable is more informative than another, and vice versa---any scenario is possible with nonzero probability.\footnote{Note this probability is well-defined because conditional entropy is a random variable.} This rules out any special condition or constraint on the marginal informativeness of any conditioning variable. In practice, it is a reasonable assumption because it is generally impossible to say with certainty that any given information source is more or less informative than any other. 

\begin{restatable}[Variables incorporated in the attention scores]{proposition}{fstprop} \label{prop1}
Assuming that all random variables are valued in standard Borel spaces and Assumption \ref{ass1} holds,
\begin{align} 
H\Big(U\mid V^{Target}\Big) &> H\Big(U\mid V^{Query}, V^{Target}\Big) \label{prop1ineq} \\
H\Big(U\mid V^{Query}\Big) &> H\Big(U\mid V^{Query}, V^{Target}\Big) \label{prop2ineq}
\end{align}
\end{restatable}
See Appendix~\ref{prop1proof} for the proof.

In particular, conditioning on $V^{Target}$ alone can only reduce the ability to correctly identify key instances. This is important because tackling MIV using naive combinations of Siamese networks and SOTA attention-based MIL models does not incorporate the information from the query into attention. In contrast, our approach uses both the query instance and the target bag to model the attention scores, and Proposition~\ref{prop1} (including Assumption \ref{ass1}) provides a theoretical foundation for it.

\subsection{MIV versus related learning problems} \label{sec:miv_concept}

\begin{figure}[t]
    \centering
    \begin{minipage}{0.75\textwidth} \includegraphics[width=\textwidth]{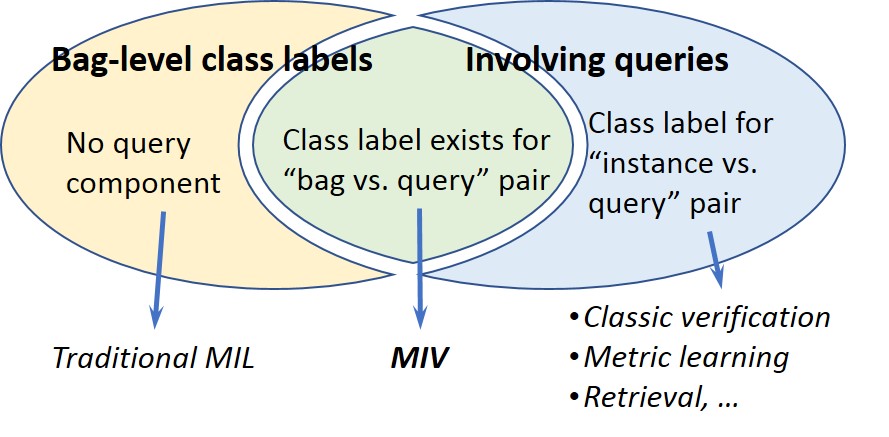}
    \end{minipage}
    \vspace{0.2cm}
    \begin{minipage}{\textwidth}
    \centering
    \begin{tabular}{@{}lc@{\extracolsep{30pt}}c@{\extracolsep{30pt}}c@{}}
        Problem& \multirow{2}{*}{MIL} & \multirow{2}{*}{MIV} & Problems\\
        characteristics& & & w/ queries\\ \toprule
        \multirow{2}{*}{Class label is for:}& \multirow{2}{*}{Bag}&\multicolumn{1}{l}{``Bag vs.}& \multicolumn{1}{l}{``Instance vs.}\\
        & &\multicolumn{1}{l}{query'' pair}&\multicolumn{1}{l}{query'' pair}\\
        Are queries involved? & No & Yes & Yes \\ \bottomrule
    \end{tabular}
    \end{minipage}
    \caption{A Venn diagram illustrating relationships between learning problems suitable for multi-instance verification (MIV), those that are suitable for traditional multi-instance learning (MIL), and learning problems that involve queries (e.g., classic verification, metric learning, standard retrieval problems, and so forth). MIV and MIL problems both require training data with bag-level labels, but differ on the existence of a query component. Classic problems involving queries, typically in the form of ``instance vs. query (instance)'' pairs, require training labels attached to these pairs of instances---i.e., at the instance-level, as opposed to the bag-level.}\label{fig:miv_venn}
\end{figure}

To highlight the significance of MIV, Figure \ref{fig:miv_venn} schematically depicts the relationship between application scenarios suitable for MIV and those corresponding to two closely related problem sets: (a) standard learning problems involving queries and (b) applications where traditional MIL is appropriate.\footnote{We thank an anonymous reviewer for suggesting the inclusion of this content to position the MIV problem.}

\subsubsection{MIV vs. standard problems involving queries}

Importantly, standard learning problems that involve queries, including metric learning, one-shot learning, and standard retrieval problems, are similar to the classic verification problem, generally \emph{not} involving the concept of ``bags of \emph{unlabeled} instances'' and bag-level class labels. Concretely, they all need training data with class labels attached to the ``instance vs. query'' pairs, where the query itself is also an instance---hence requiring the ground truth of the instance pairs during training. For example, in the standard retrieval problem (likewise in metric or one-shot learning), a ``match'' can only be defined if a unit of the searched content can be labeled as ``matched''(positive) or ``not matched''(negative) with the query, requiring such labels of all instance pairs in the training data. The labels may be explicit (by annotation) or implicit (inferred from the searched content's location, information sources, etc.), as commonly seen in standard ``question answering''  or ``image retrieval'' tasks. Even in self-supervised training, labels of instance pairs are still required---for example, a label indicating whether an instance is (a distortion of) the query is typically constructed for self-supervised metric learning.

The availability of ground-truth labels for instance pairs can be viewed as an instance-level label requirement, which is in stark contrast to the bag-level label requirement of MIV---as stated in Lemma~\ref{lemma1} (Section~\ref{def_miv}), an MIV exemplar can be represented as a bag of \emph{unlabeled} instance pairs, with only a bag-level label available. In practical applications, if the instance-level label requirement is met, the MIV formulation is \emph{not} needed. Therefore, in terms of practical applications, MIV and standard problems involving queries are applicable to mutually exclusive scenarios, and the appropriate formulation should be adopted depending on the available data.

It is also worth noting that even if instance-level labels exist, it may be appropriate to conceptually view the problem at hand as an MIV task with bag-level labels when annotation errors, embedding quality issues, or constraints from feature extractors render individual instance-level labels unreliable. 

\subsubsection{MIV vs. MIL}

\begin{table}[t]
    \centering
    \begin{tabular}{@{}llcl@{}} 
        \multirow{2}{*}{\textbf{Approaches}}& \multicolumn{1}{c}{\textbf{Motivated by what}}& &\multicolumn{1}{c}{\textbf{Novelty in relation to}}\\ 
        & \multicolumn{1}{c}{\textbf{representation of MIV data?}} & &\multicolumn{1}{c}{\textbf{traditional MIL}}\\ \toprule
        \multirow{3}{*}{CAP (ours)} &A bag of (unlabeled) instances& &\emph{Deviation} from traditional MIL\\
        &plus a query \emph{related} to the bag&&due to additional role of query;\\ 
        &(satisfying Assumption~\ref{ass1})&&\emph{New} MI pooling algorithm\\ \midrule
        Baseline & A bag containing (unlabeled)& &\emph{New}: handling instance-pairs;\\
        model&pairs of instances (Lemma~\ref{lemma1})& &\emph{Not new}: aggregation to bag-level;\\ \midrule
        Benchmark& A bag plus a query independent of& &\emph{No novelty}: naive adaptation of\\
        models&the bag (violating Assumption~\ref{ass1})&&existing MIL algorithms;\\ \bottomrule
    \end{tabular}
    \caption{A summary of approaches considered in our experiments, including their conceptual motivations based on different data representations (or assumptions), and their levels of novelty in relation to the existing MIL literature. Both baseline and benchmark models are motivated by representations more akin to that of traditional MIL, so as to leverage existing MIL algorithms, either partially or fully. In contrast, our approach (CAP) deviates from MIL by focusing on the query's role in inducing bag-level embeddings, in addition to its classic role in verification (as an ``anchor'').}
    \label{tab:miv_approaches}
\end{table}

Compared to traditional MIL, MIV  exhibits additional complexity and is independently significant. The difference between MIV and MIL rests on the existence of a query component in the feature space. The query is subject to transformations (or assumptions), making MIV a multi-faceted problem comprising multiple representations of the data. Table~\ref{tab:miv_approaches} summarizes three different representations of the data in MIV problems (middle column), which correspond to different approaches considered in our study (leftmost column) and exhibit different levels of novelty, or deviation, in relation to standard MIL (rightmost column).

The benchmark models we consider (in Section~\ref{sec:bmodels}) are motivated by the assumption that the query is independent of the target bag---implying \emph{zero} probability that the query can be more informative than instances of the target bag---violating Assumption~\ref{ass1} in Section~\ref{sec:as}. This representation makes MIV a trivial extension of MIL, and the query only has its classic role in verification as an ``anchor''. Consequently, all benchmark models can naively apply existing MIL algorithms to construct bag-level embeddings without considering the query. 

The second representation, giving rise to our baseline model (\emph{cf.} Section~\ref{sec:baseline}), is that from Lemma~\ref{lemma1} in Section~\ref{def_miv}, which is based on bags of instance-pairs. The baseline model handles instance-pairs using Siamese networks and processes bags of instance-pairs in an MIL manner. It is partially novel compared to standard MIL approaches because there is no dedicated mechanism in standard MIL to deal with instance pairs. 

In contrast to the baseline and benchmark models, our new CAP-based approach is motivated by a representation satisfying Assumption~\ref{ass1}, and consistent with Proposition~\ref{prop1}. It focuses on the query's additional role in inducing bag-level embeddings, resulting in the novel MI pooling algorithm, CAP (see Section~\ref{sec:cap}). Notably, this perspective on the role of the query, and the resultant pooling mechanism, do not exist in the MIL literature, rendering MIV distinct from MIL.

Considering the approaches in Table~\ref{tab:miv_approaches}, our empirical results show that the more a representation deviates from that of MIL, the more empirical success we can obtain in MIV applications, signifying the importance of choosing an approach based on suitable representations or assumptions---in our paper, it should be Assumption \ref{ass1}, Proposition \ref{prop1}, and proper consideration of the query’s role in inducing bag-level embeddings.

\section{Model architectures} \label{sec:arch}
We now describe the different modeling approaches compared in our experiments in detail, including the baseline model, the benchmark models, and the new approach based on cross-attention pooling (CAP), instantiated with two novel attention functions we propose.

All models compared in our study share the same architecture for the first two layers and the final two layers of the neural networks. In the first two layers, we obtain fixed-length feature vectors by embedding the query instance and all instances in the target bag using appropriate feature extractors from the CV or NLP domains, as required by the application at hand, followed by a layer normalization (LayerNorm,~\cite{layernorm16}). We denote the feature vectors \emph{after} LayerNorm as $v^{query}\in \mathbb{R}^{1\times C}$ and $v^{target}\in \mathbb{R}^{N\times C}$ respectively, where $N$ is the number of instances in a bag (``bag size''), and $C$ is the number of channels. Note that bag size $N$ varies across exemplars.
For the final layer, regardless of architecture, we adopt a binary cross-entropy loss, in which each exemplar's prediction is $p=sigmoid(sim)$, where $sim$ is a Siamese-twin similarity score:
\begin{align}\label{gen_sim}
sim = \sum_{i=1}^C \alpha_i f(v^P_i, v^Q_i),
\end{align}
where $\alpha\in\mathbb{R}^{1\times C}$ is a vector of learnable parameters, and $v^Q, v^P\in \mathbb{R}^{1\times C}$ are the two outputs of the Siamese twins, corresponding to (a) the encoded query instance and (b) the bag-level representation, which is the attention-weighted sum of the encoded target-bag instances, respectively (\emph{cf.} Figure \ref{fig:cap_arch}). The function $f(v^P_i, v^Q_i)$ measures the similarity between $v^P$ and $ v^Q$. Equation~\eqref{gen_sim} is the same as that used by \cite{siamese15} except that we use the product function, $f(v^P_i, v^Q_i) = v^P_iv^Q_i$ instead of $L_1$ distance, which yields
\begin{align} \label{siamese_sim}
sim = v^Q diag(\alpha) \big(v^P\big)^T,
\end{align} 
where $diag(\alpha)$ denotes a $(C\times C)$ diagonal-matrix based on $\alpha$.

\subsection{Our approach: Cross-Attention Pooling (CAP)} \label{sec:cap}

\begin{figure}[!ht]
     \centering
     \begin{subfigure}[c]{.96\textwidth}
         \includegraphics[width=\textwidth]{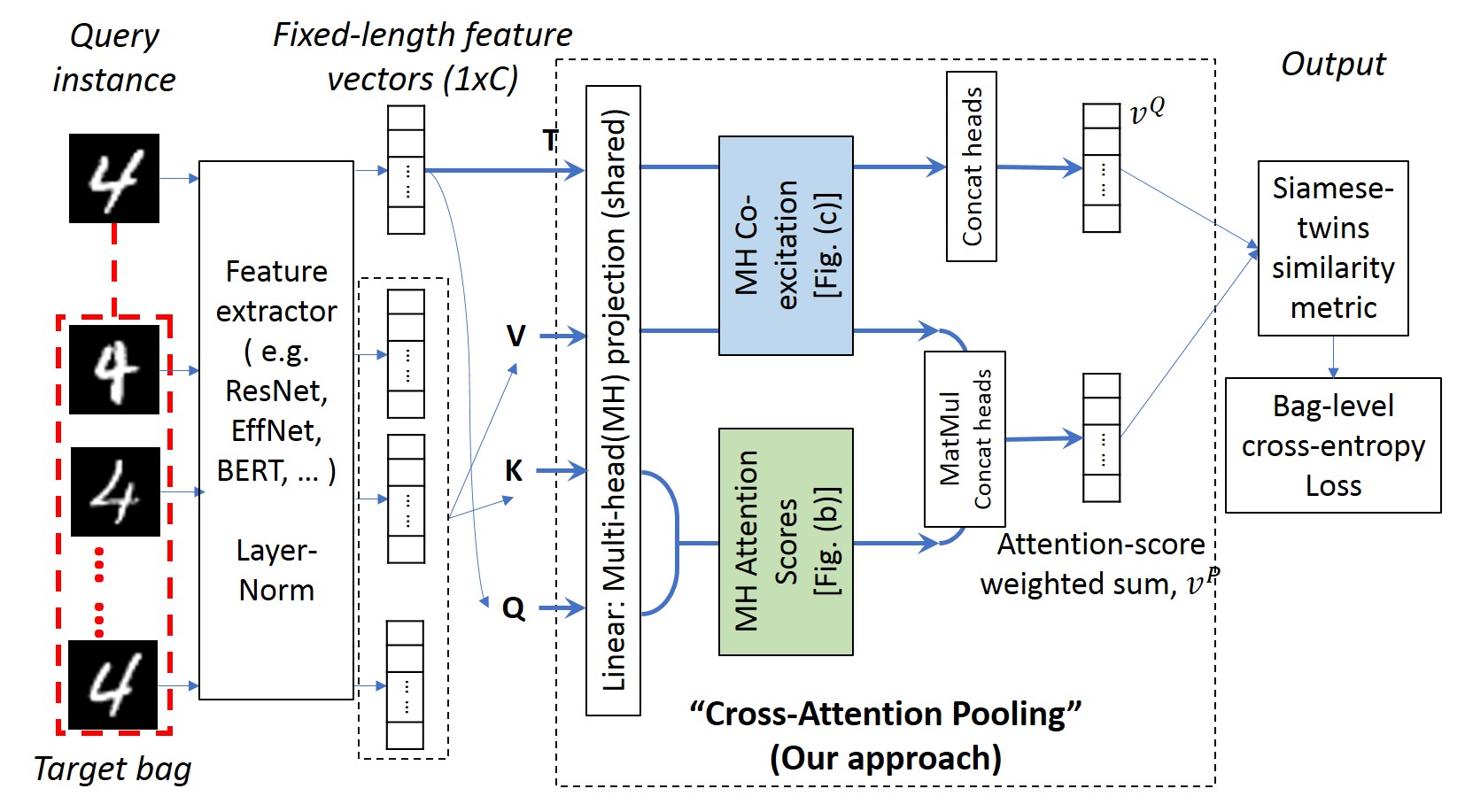}
         \caption{CAP architecture} \label{architecture_a}
     \end{subfigure}
     \begin{subfigure}[c]{0.54\textwidth} 
         \includegraphics[width=\textwidth]{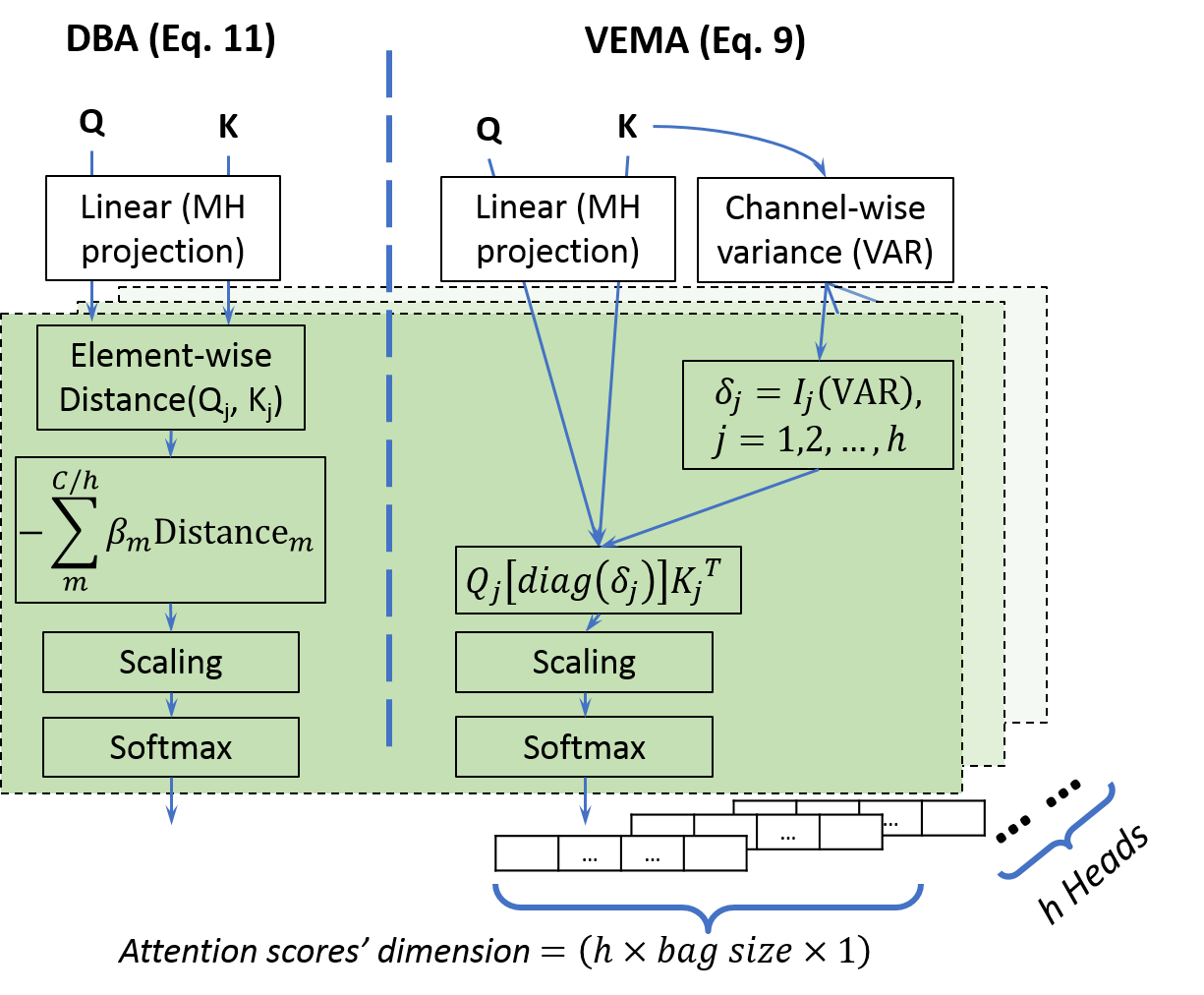}
         \caption{Multi-head attention scores}\label{architecture_b}
     \end{subfigure}
     \begin{subfigure}[c]{0.45\textwidth} 
         \includegraphics[width=\textwidth]{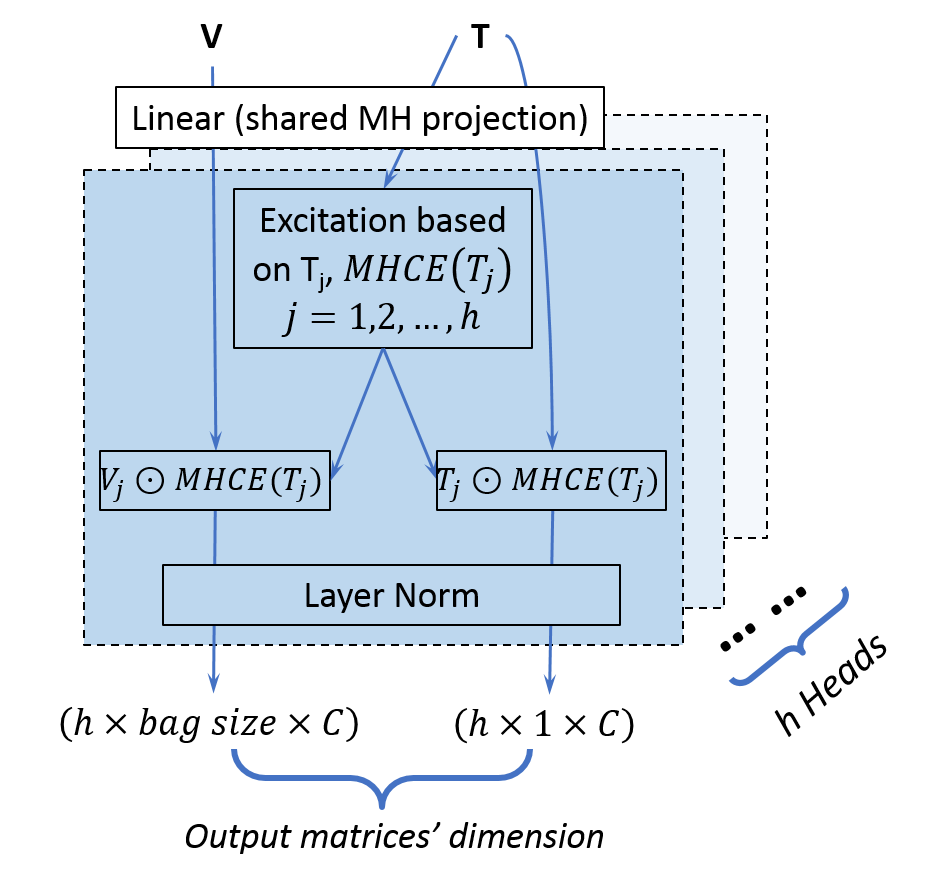}
         \caption{Multi-head co-excitation} \label{architecture_c}
     \end{subfigure}
     \caption{Model architecture for ``Cross-Attention Pooling'', shown in (\ref{architecture_a}), and its key components: (\ref{architecture_b}) multi-head attention scores and (\ref{architecture_c}) multi-head co-excitation, best viewed in colour. In (\ref{architecture_a}), \textbf{Q}, \textbf{K}, \textbf{V}, \textbf{T} denote the vectors of ``query'', ``key'', ``value'' and ``transformed query instance'' (to be transformed into $v^Q$) , respectively. In (\ref{architecture_b}), $diag(\delta)$ denotes a $(D\times D)$ diagonal-matrix based on $\delta$ with the function $I_j(\cdot)$ defined by Equation~\eqref{vema_delta}, \textbf{Eq. 9} and \textbf{Eq.11} denote Equations~\eqref{vema} and \eqref{dba} respectively. In (\ref{architecture_c}), $\odot$ denotes element-wise multiplication, and the function $MHCE(\cdot)$ is defined by Equation~\eqref{ce}.}
     \label{fig:cap_arch}
\end{figure}

The main difference between the algorithms investigated in this study is how they conduct MI-pooling of $v^{target}$ to get $v^P$. Our new pooling method CAP is motivated by two considerations. First, as implied by Proposition \ref{prop1}, it is \emph{strictly} better to incorporate $v^{query}$ to model attention scores for the pooling of $v^{target}$. To operationalize this idea,  we construct the attention score for each instance in the target bag ($v^{target}$) by attending it to $v^{query}$ as a cross-reference, inspired by the  ``query-key-value'' framework in the transformer decoder (\cite{transformer17}). 

The second consideration comes from the challenge of handling tasks in which all instances in a bag look alike, exemplified by the QMNIST task in our study. 
The challenge is how to make key instance(s) stand out among similarly looking instances, through an uneven distribution of their attention scores. We tackle this using two strategies. One is to create new functions that provide unevenly distributed attention scores. Another is to excite channels of the ``value''-vector. The former gives rise to our novel attention functions, namely variance-excited multiplicative attention (VEMA) and distance-based attention (DBA). The latter leads to our multi-head adaptation of ``co-excitation'' (\cite{sce19}), a variant of the ``squeeze and excitation'' network (\cite{se18}). 

Figure \ref{fig:cap_arch} depicts the entire Siamese network architecture, including our CAP approach highlighted by the dotted rectangle (Figure~\ref{architecture_a}), together with its key components for computing attention scores (Figure~\ref{architecture_b}) and ``co-excitation'' (Figure~\ref{architecture_c}) highlighted by green and blue colours respectively---those components are explained more formally in Section \ref{sec:cap_comp} below. As shown in the figure, there are two outputs of CAP: one is the bag-level representation $v^P$, another is the transformed query instance $v^Q$. $v^P$ is a function of the triad of ``query, key, value'' denoted by ``\textbf{Q}, \textbf{K}, \textbf{V}'', whereas $v^Q$ is a function of $v^{query}$ denoted by ``\textbf{T}'' in the figure. The functions producing $v^P$ and $v^Q$ are formally specified by Equation~\eqref{perhead}.

\subsubsection{Multi-head Cross-Attention Pooling} \label{sec:mhcap}

Letting $\mathbf{Q}=v^{query}$, $\mathbf{K}=\mathbf{V}=v^{target}$, $as(\cdot,\cdot)$ be the attention-score function, $LN(\cdot)$ be layer normalization, $MHCE(\cdot)$ be ``multi-head co-excitation'', we model the outputs of the Siamese twins as
\begin{subequations} \label{perhead}
\begin{align}
v^P &= concat(O^P_1, O^P_2, \dots, O^P_h),\qquad v^Q = concat(O^Q_1, O^Q_2, \dots, O^Q_h), \nonumber \\
O^P_j 
&= MatMul\Bigg[\underbrace{as_j\Big(v^{query}W_j, v^{target}W_j\Big)}_{1\times N}, \underbrace{LN_j\Big(v^{target}W_j \odot MHCE_j(v^{query}) \Big)}_{N\times D}\Bigg], \label{perhead_p} \\ 
O^Q_j &= \underbrace{LN_j\Big(v^{query}W_j \odot MHCE_j(v^{query}) \Big)}_{1\times D}, \label{perhead_q} \qquad j = 1,2,\dots,h
\end{align}
\end{subequations}
where $h$ is the number of heads, $D=\frac{C}{h}$, $W_j \in \mathbb{R}^{C\times D}$ are learnable weights of the multi-head linear projections shared by $\mathbf{Q},\mathbf{K},\mathbf{V}$, and the subscript $j$ of $as_j(\cdot,\cdot)$, $MHCE_j(\cdot)$, $LN_j(\cdot)$ indicates the $j^{th}$ head. $MatMul$ and $\odot$ denote matrix and element-wise multiplication. 




Next, we provide details of the functions $as(\cdot,\cdot)$ and $MHCE(\cdot)$ for an individual head within the multi-head setting, omitting the subscript $j$ for brevity.

\subsubsection{Components of Cross-Attention Pooling} \label{sec:cap_comp}

VEMA is one of the two attention functions proposed in this paper. The rationale is to construct a suitable variant of the ``multiplicative attention'' function popularized by the transformer (\cite{transformer17}). 
To this end, VEMA creates an adjustment using channel-wise variance (across instances within a bag). Intuitively, when all instances in a bag look similar to each other, channels that make them \emph{more different}, i.e., channels with \emph{higher variance}, may make it easier to identify key instances. Accordingly, we design the ``excitation by variance'' of channels, which can be viewed as a feature selection (or channel selection) mechanism in attention. We expect it to produce more unevenly distributed attention scores, and hence more accurate predictions of the key instances, a hypothesis to be tested by the experiments in Section \ref{sec:experiments}. Formally, we have
\begin{equation}\label{vema}
as(Y, Z) = softmax\Big(\frac{Y diag(\delta)Z^T}{\sqrt{D}}\Big),
\end{equation}
where $Y$ and $Z$ have shape $(1\times D)$ and $(N\times D)$ respectively, and $\delta\in\mathbb{R}^{1\times D}$. $\delta$ is modeled as a gating mechanism, 
\begin{equation}\label{vema_delta}
\delta = sigmoid\Big[ relu\Big((variance(v^{target})-1)R\Big)S \Big], 
\end{equation}
where $variance(v^{target}) \in \mathbb{R}^{1\times C}$ denotes the channel-wise variance of $v^{target}$, $R\in\mathbb{R}^{C\times C}$ and $S\in\mathbb{R}^{C\times D}$ are learnable parameters (with $S$ projecting channels to a single head), and $sigmoid$ and $relu$ are element-wise activation functions.

Our motivation for the second attention function DBA is to establish an equally effective but simpler alternative to VEMA. Considering multiplicative attention as a generalized ``cosine similarity'', we surveyed alternative similarity measures, which led to DBA as a new attention function. More precisely, the attention scores of DBA are driven by the negative sum of weighted distances between corresponding channels of $\mathbf{K}$ and $\mathbf{Q}$. In this paper, we focus on $L_1$-distance, although other distance metrics are also possible.

The DBA attention-score function is
\begin{equation}\label{dba}
as(Y, Z) = \underset{n\in\{1,\dots,N\}}{softmax}\Big(\frac{c - \sum_m^D \beta_m|Y_{0,n,m} - Z_{0,n,m}|}{s}\Big),
\end{equation}
where $Y$ and $Z$ are broadcast to the same shape $(1\times N\times D)$, $|\cdot|$ yields element-wise absolute value, $\beta \in \mathbb{R}^{1\times D}$ is a vector of learnable parameters for a single head (with different $\beta$ for different heads), and $c=\sqrt{\frac{4}{\pi}}D$ and $s=\sqrt{(2-\frac{4}{\pi})D}$ are two constant scalars, broadcast as $(1\times N)$ to be compatible with Equation~\eqref{dba}. Like the scaling factor in the transformer (\cite{transformer17}), $c$ and $s$ are normalizing factors to maintain numerical stability.\footnote{$c$, $s$ are  the mean and standard deviation of $\sum_j^D|a_j-b_j|$ assuming all $a_j, b_j$ are i.i.d. $\sim \mathcal{N}(0,1)$.} 

To elucidate the rationale behind DBA, we note that distance metrics can be used to form variation measures (or measures of dispersion). To see this, if we average,  across all instances in a bag, the channel-wise distances between the bag ($\mathbf{K}$) and the query ($\mathbf{Q}$), we obtain a measure of ``$\mathbf{K}$'s variation around $\mathbf{Q}$'' for each channel. For example, we get ``mean absolute deviation around $\mathbf{Q}$'' if we use $L_1$-distance, and ``variance around $\mathbf{Q}$'' if we use $L_2$-distance. Hence, channels with \emph{higher variation} around $\mathbf{Q}$ automatically contribute \emph{more} to the attention score in Equation~\eqref{dba}, because high variation implies greater distance. Therefore, DBA is similar in spirit to VEMA but provides a simplification---comparing Equation~\eqref{dba} to \eqref{vema}, we can see DBA contains no explicit formula of variation, even though it implicitly achieves the same effect, and with fewer learnable parameters ($C$ vs. $2C^2$).\footnote{The actual implementation of VEMA also includes a bias term, resulting in $2(C^2+C)$ parameters.} Geometrically, distances are also more robust to changes of data origin, than the (non-centered) cosine similarity. The efficacy of this simplification and its robustness, reflected in classification and key instance detection accuracy, is to be examined in Section \ref{sec:experiments}.

Finally, for the MHCE module, the ``excitation'' function (\cite{se18}) is a gating mechanism akin to the ``gated linear unit''(\cite{glu16}), which is well-known for its ability to enable feature selection. Co-excitation is to apply the same ``excitation'' function driven by the query to both Siamese twins---that is, to activate simultaneous selection of features for both query and target bag. More concretely, an ``excitation'' function is shared by both Equations~\eqref{perhead_p} and \eqref{perhead_q}, \emph{prior to} element-wise multiplications with $(v^{target}W_j)$ and $(v^{query}W_j)$ respectively. We adopt ``co-excitation'' in a multi-head context, thus naming it ``MHCE''. Formally, MHCE's ``excitation'' function is: 
\begin{equation}\label{ce}
MHCE(x) = sigmoid\Big[relu\big(xJ\big)M\Big],
\end{equation}
where $J\in\mathbb{R}^{C\times C}$ and $M\in\mathbb{R}^{C\times D}$ are learnable parameters, $sigmoid$ and $relu$ are element-wise activation functions, and $x\in \mathbb{R}^{1\times C}$. 
We emphasize that the same $x=v^{query}$, as well as the same learnable parameters, in $MHCE(x)$ are shared between Equations~\eqref{perhead_p} and \eqref{perhead_q}. Note that $MHCE(\cdot)$ is multi-headed, because $M$ projects all the channels to a head.


\subsection{Baseline and benchmark models}\label{sec:bmodels}
\begin{figure}[t]
     \centering
     \begin{subfigure}[b]{0.45\textwidth}
         \centering
         \includegraphics[width=0.85\textwidth]{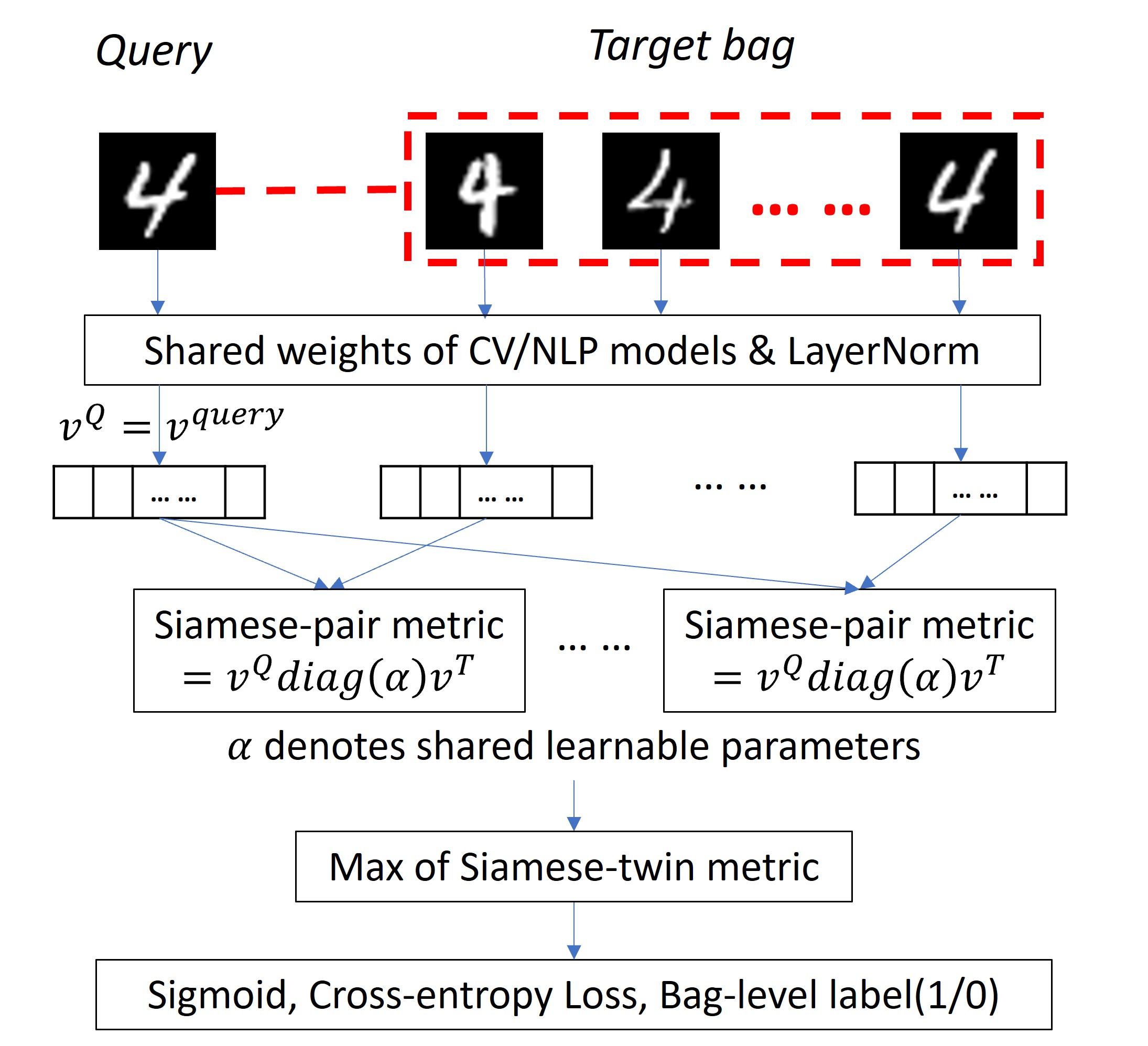}
         \caption{Baseline approach}
         \label{arch_a}
         \vspace*{9pt}
     \end{subfigure}
     \hfill
     \begin{subfigure}[b]{0.45\textwidth}
        \centering
        \includegraphics[width=0.75\textwidth]{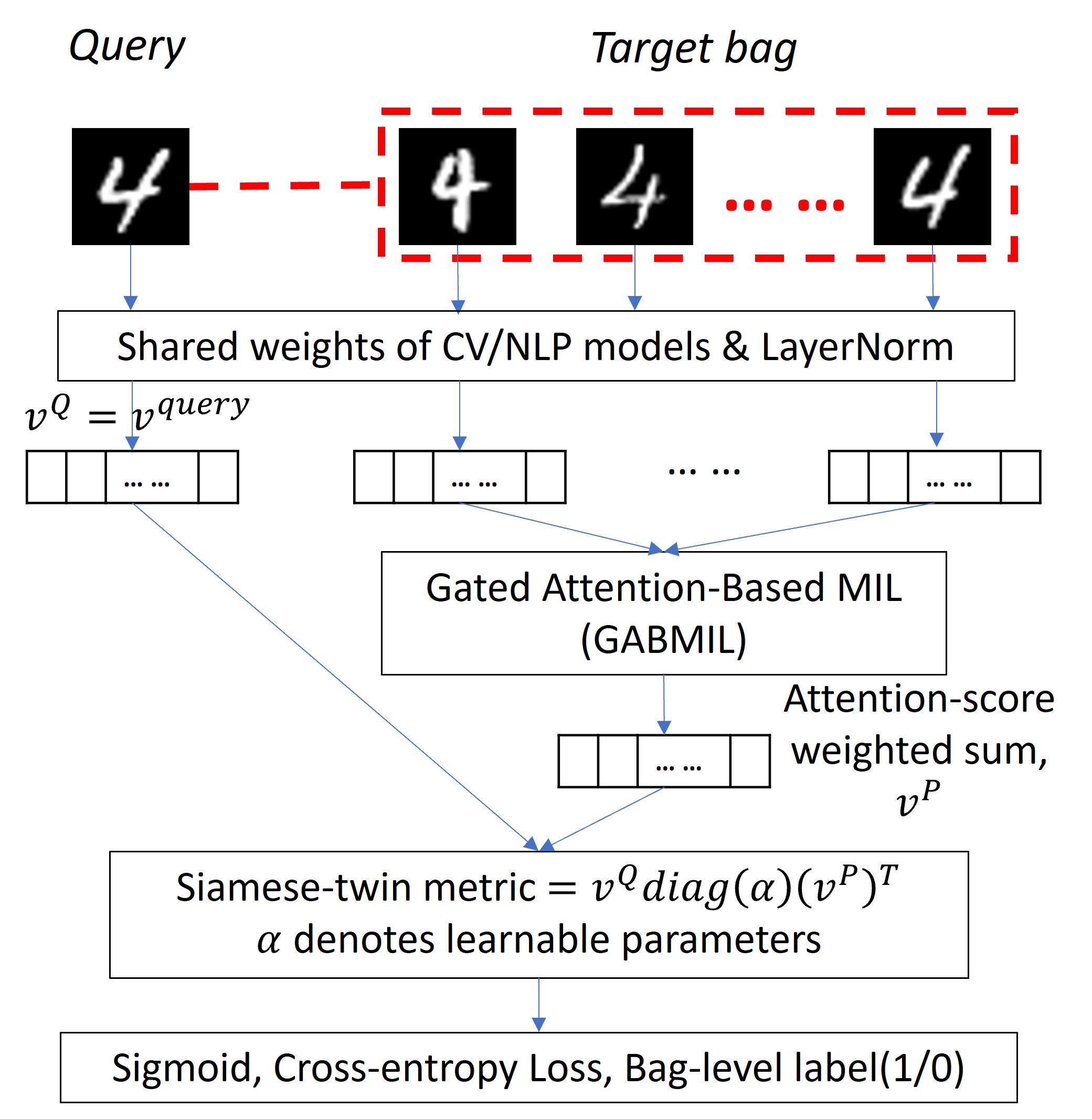}
        \caption{Approach using GABMIL}
        \label{arch_b}
        \vspace*{9pt}
     \end{subfigure}
     \begin{subfigure}[b]{0.45\textwidth}
         \centering
         \includegraphics[width=\textwidth]{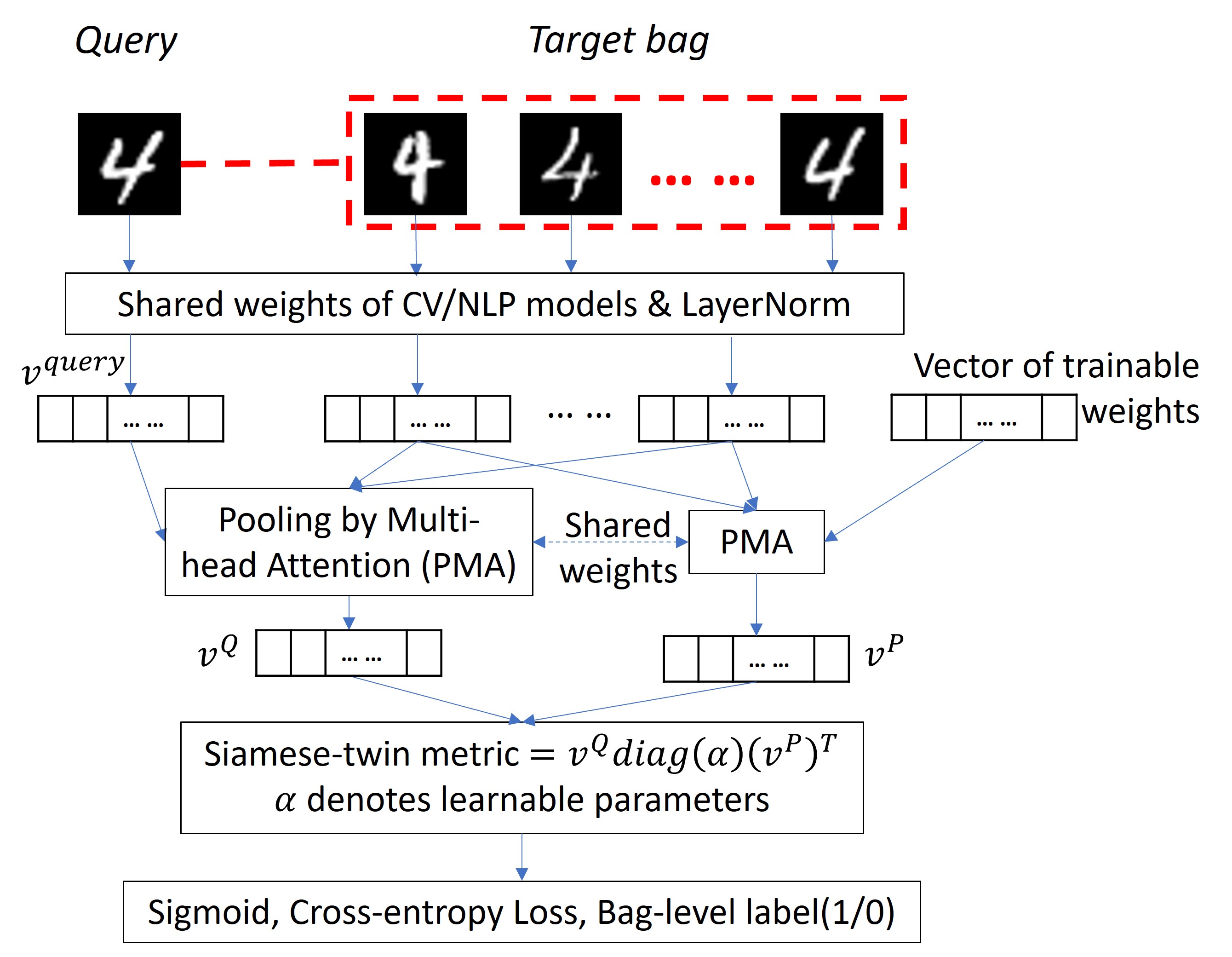}
         \caption{Approach using PMA}
         \label{arch_c}
     \end{subfigure}
     \hfill
     \begin{subfigure}[b]{0.45\textwidth}
         \centering
         \includegraphics[width=\textwidth]{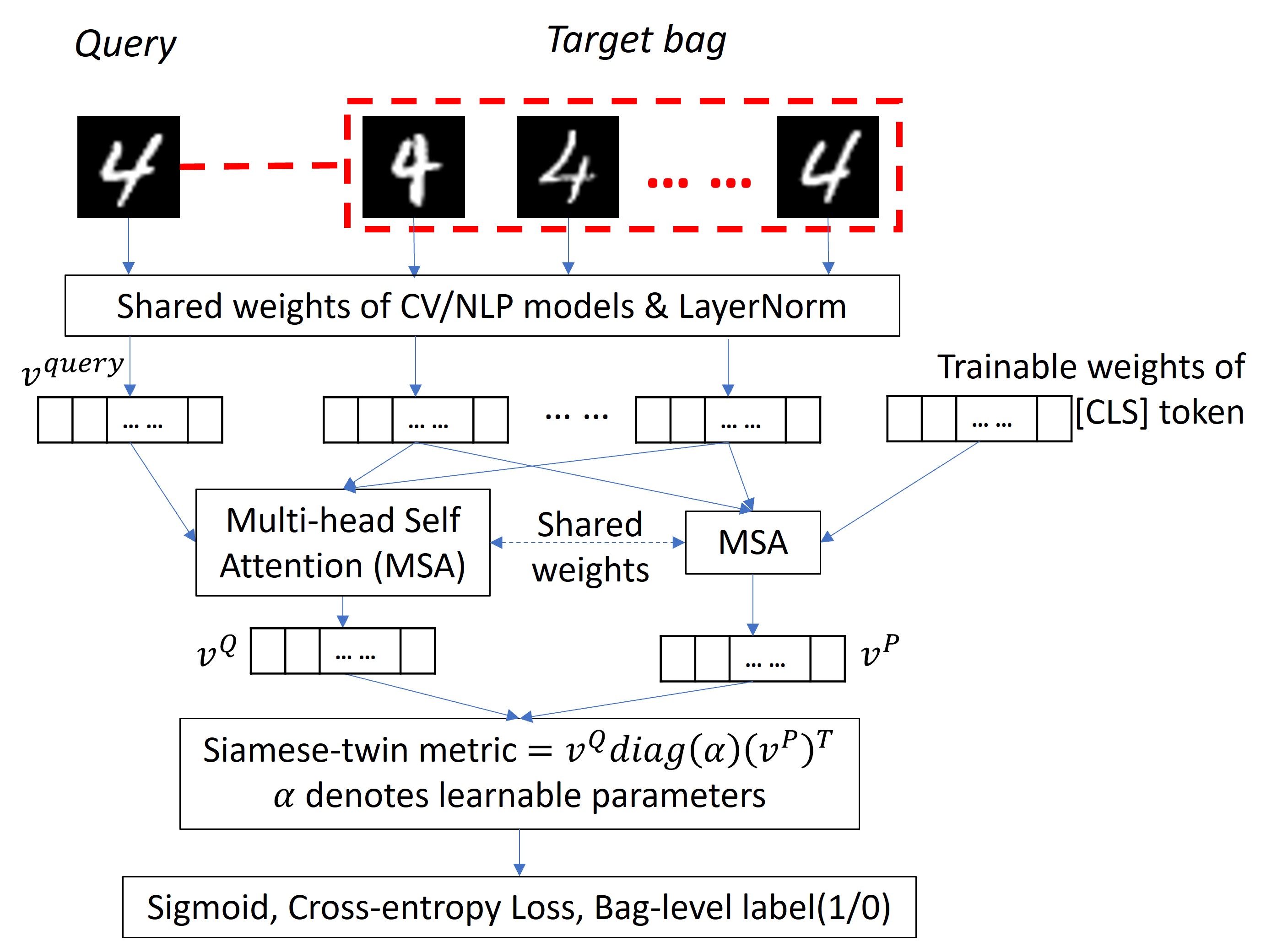}
         \caption{Approach using MSA}
         \label{arch_d}
     \end{subfigure}
     
     \caption{Model architectures of the baseline and the benchmarks}
     \label{fig:arch_bm}
\end{figure}

To provide a baseline for all models, we create a simple model for MIV, inspired by Lemma~\ref{lemma1} (Section~\ref{def_miv}). We also adapt SOTA methods from the  MIL literature to develop three benchmark models that enable model comparison. The three SOTA methods are: ``Gated-Attention Based MIL'' (GABMIL) from~\cite{gabmil18}, ``Pooling by Multi-head Attention'' (PMA) from~\cite{pma19}, and ``Multi-head Self Attention'' (MSA) used in~\cite{transmil21}. While the benchmark models are all attention-based, we also experiment with some non-attention-based MI-pooling methods in Section \ref{sec:experiments}, namely MI-Net from~\cite{minet18} and bi-direction LSTM from~\cite{mil-lstm20}. The architectures of the baseline and attention-based benchmark models are depicted in Figure \ref{fig:arch_bm} and explained in the following. The architectures of non-attention-based models are similar to that of the benchmark using GABMIL (Figure \ref{arch_b})---that is, $v^P$, the non-attention-based bag-level representation of $v^{target}$, and $v^Q = v^{query}$ are the two outputs of the Siamese twins---and are omitted here for brevity.

\subsubsection{The baseline} \label{sec:baseline}
The baseline model simply pairs $v^{query}$ with each of the $N$ vectors comprising $v^{target}$ and computes the Siamese-twin similarity metric, in Equation~\eqref{siamese_sim}, for each pair. It then takes the maximum of the $N$ similarity scores as the logit of this exemplar, to be fed into the activation and loss function, see Figure \ref{arch_a}.

Because the baseline model aggregates a bag based on its elements' scores without an explicit bag-level representation, it is analogous to an extension of instance-level MIL methods, e.g., mi-Net of \cite{minet18}. Implicitly, it nevertheless still has a bag-level representation for the target bag, which is the feature vector of the instance whose corresponding pair attains the maximum score. More precisely, the baseline can be viewed as a very simple attention model, since taking the maximum implements an attention mechanism using the $\mathbbm{1}_{argmax}$ function instead of the $softmax$ function used by other models, where $\mathbbm{1}_{argmax}$ denotes an indicator function providing a vector based on the set of indices returned from the $argmax$ function.\footnote{In case there is more than one maximum, the output of $\mathbbm{1}_{argmax}$ is normalized by $L_1$-normalization to sum to 1, i.e., $\frac{|x_n|}{\sum_n|x_n|}$.} Like in other attention-based methods, the (implicit) bag-level representation in the baseline model is also the attention-score weighted sum of the instances' representations:
\begin{align*}
v^P = \mathbbm{1}_{argmax}\big(v^{query} diag(\alpha) (v^{target})^T\big) \cdot v^{target},
\end{align*}
where $\alpha$ is the learnable parameter vector in Equation~\eqref{siamese_sim}.


This baseline model is limited, due to (1) its simplified architecture with very few learnable parameters, (2) the $argmax$ attention function that allows only those instances that yield maxima to represent the entire bag, and (3) parameters $\alpha$ that are shared by both the Siamese-twin metric and the attention mechanism. Nonetheless, this simple model explicitly handles bags of paired instances and captures the essence of a representation of MIV data in Lemma~\ref{lemma1} (Section~\ref{def_miv}). Hence, this model is a useful baseline for model comparison.

\subsubsection{Benchmark using GABMIL}
As the first benchmark, we adopt the MI pooling method of Gated-Attention-Based MIL (GABMIL) from \cite{gabmil18} to aggregate $v^{target}$, which computes one of the Siamese twins $v^P$ as the attention-weighted sum of $v^{target}$, and another as $v^Q = v^{query}$ (Figure \ref{arch_b}).

\subsubsection{Benchmark using PMA}\label{sec:pma}
The second benchmark applies an MI pooling method based on the transformer decoder, called ``Pooling by Multi-head Attention'' (PMA) by \cite{pma19}. PMA has the same architecture as the transformer decoder (without positional encoding and dropout), used to aggregate $v^{target}$. \cite{pma19} define $PMA_k$ by applying the transformer's multi-head cross-attention on a learnable set of $k$ seed vectors $\in \mathbb{R}^{k\times C}$. 

We initially used $PMA_1$ to aggregate $v^{target}$: $v^P = PMA_1(v^{target}; P)$, $v^Q = v^{query}$, where $P\in \mathbb{R}^{1\times C}$ is a learnable seed vector. However, exploiting the Siamese-twin structure, we found it useful to compute $v^Q$ as an aggregation of $v^{target}$ by treating $v^{query}$ as a seed vector. More precisely, we stack $v^{query}$ on $P$, and compute the outputs of the Siamese twins as follows (see also Figure \ref{arch_c}):  
\[
(v^P, v^Q) = PMA_2\Big(v^{target}; (P, v^{query})\Big).
\]
We found that $PMA_2$ yields slightly better classification performance than $PMA_1$. Hence, we adopt $PMA_2$ as the benchmark in the experiments presented in our study.

Notably, $PMA_2$ consists of paired PMAs that adapt to Siamese networks to solve a verification problem, where PMA is constructed using the transformer's cross-attention mechanism. In this sense, this benchmark can be seen as an adaptation of standard, ``transformer-style'' cross attention---without fundamentally changing its core architecture---to the MIV setting.

\subsubsection{Benchmark using MSA}
Finally, there is an extensive range of applications and variations of transformer encoders in the literature (e.g., \cite{transmil21, vit21}). The key component of these methods is multi-head self-attention (MSA), characterized by the full self-attention of all the instances within a bag to themselves. 

MSA per se cannot aggregate multiple instances because it retains the same positions in the output sequence as in the input sequence. To use MSA as an MI pooling method, the common practice is to include an additional, special token, ``[CLS]'', in the input sequence, whose embedding vector, $v^{[CLS]}$, is learnable. One can treat the vector at the position of ``[CLS]'' in the output sequence as the aggregation of the rest of the inputs.

Analogously to $PMA_2$, we treat $v^{query}$ as a feature vector of another special token. We then apply the same MSA (without positional encoding and dropout) independently to both $v^{[CLS]}$ and $v^{query}$. This gives rise to the Siamese twins (in Figure \ref{arch_d}) as
\[
(v^P, v^Q) = MSA\Big(v^{target}; (v^{[CLS]}, v^{query})\Big).
\]

Following common practice to stack multiple MSA layers when aggregating $v^{target}$ (see, e.g., \cite{transmil21}), we stacked two layers of MSA---adding more layers did not appear to add value---in our experiments.

We highlight that none of the three benchmark methods has a natural mechanism to incorporate $v^{query}$ as part of attention when producing the bag-level representation $v^P$. Although the Siamese-twin similarity metric may induce a reference to $v^{query}$ in the attention-weighted aggregation of $v^{target}$, this relationship is not modeled explicitly, which may make it less effective for the identification of key instances and for classification. Section \ref{sec:experiments} provides empirical evidence that supports this hypothesis: the benchmarks are not better, and sometimes significantly worse, than even the simple baseline.

\section{Evaluating performance of key instance detection} \label{sec:explainability}

Attention-based MIL models, for example, the models presented in (\cite{gabmil18, transmil21}), reflect the importance of the ability to identify ``key instances'' using attention scores extracted from models. Following this work, and similar other studies (including~\cite{explain21, milinterpret22, wsol23}) we quantitatively evaluate the ability to identify key instances using attention scores, in the context of MIV. This is possible when the ground truth of key instances is available in the datasets, as is the case in our study.

\begin{figure}[t]
    \centering
    \includegraphics[width=0.9\textwidth]{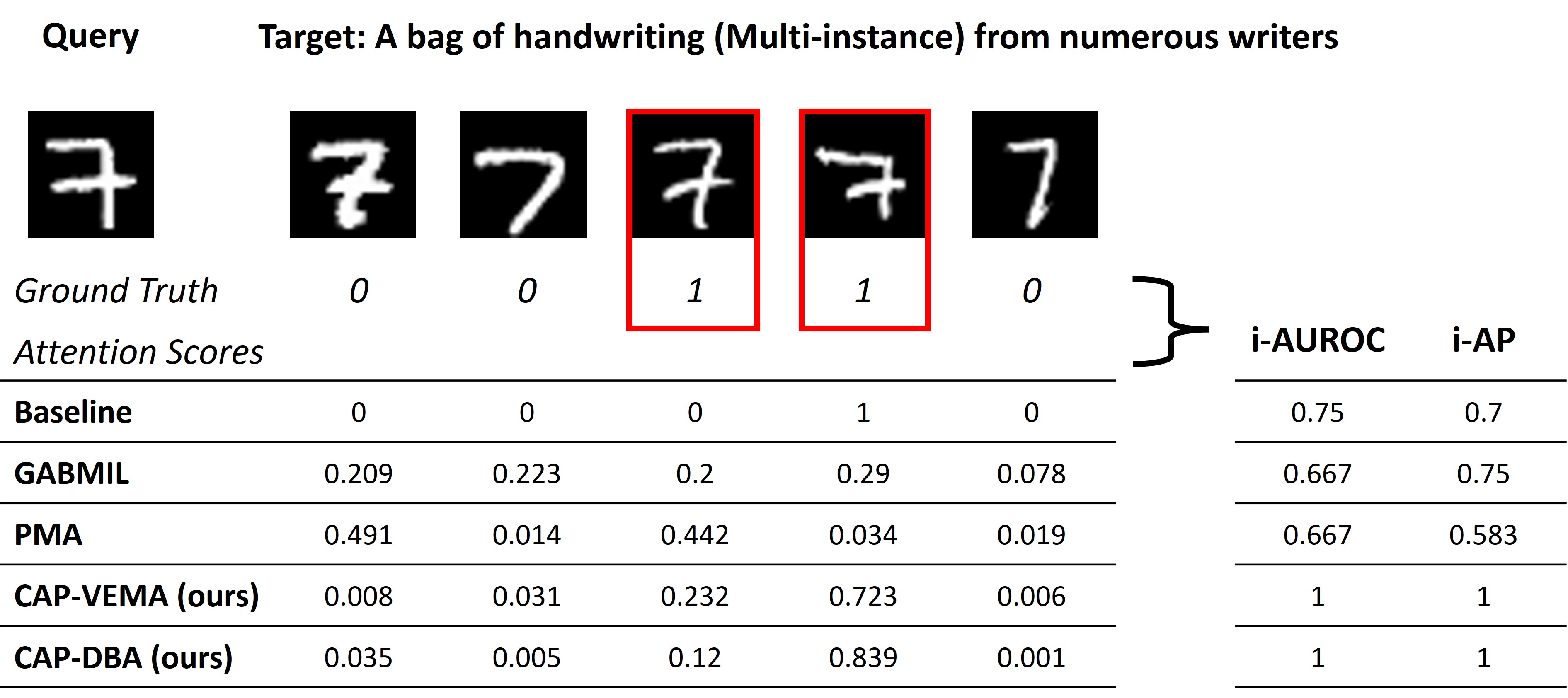} 
    \caption{An illustration of the quantitative evaluation of key instance detection. Note that ground-truth key instances, highlighted by red rectangles, are unavailable during training.}
    \label{fig:attn_quality}
\end{figure}

More precisely, for each positive exemplar in the test dataset, we match the ground truth of its key instances' identifiers to a model's attention scores. When there are multiple heads, we follow standard practices by averaging the attention scores across all heads to identify the key instances. Figure \ref{fig:attn_quality} provides an illustration of such matching for a test QMNIST exemplar. This process is related to evaluation for ``weakly-supervised object localization'' in CV domains (\cite{explain21, wsol23}), from which we employ two \emph{threshold-independent} ranking metrics, namely pixel ``area under ROC curve'' (AUROC), and pixel ``average precision'' (AP). 

The threshold independence comes from the definitions of AUROC and AP---they are summary statistics of ROC and PR curves (see, e.g., \cite{prcauc06}), on which each point evaluates the classification performance based on a single decision threshold. Consequently, by summarizing entire curves, both metrics measure the performance across all possible classification thresholds and thus are \emph{threshold-independent}. The reasons to require threshold independence in our case are twofold: one, to avoid controversies regarding how thresholds are determined to predict key instances (see, for example, \cite{wsol23}); two, for attention-based models, selecting thresholds may be less straightforward because attention scores must sum to $1$, and the bag size $N$ varies across exemplars. Therefore, sensible thresholds may need to be dynamic for exemplars with varying sizes, a subject that is beyond the scope of this paper.

We adopt the same computation as pixel AUROC and pixel AP, except that ``pixel'' becomes ``instance'' in our case. Accordingly, we call them instance AUROC (i-AUROC) and instance AP (i-AP), both of which assess key instance detection performance for a single exemplar. To summarize the two metrics for the entire test dataset, we take the average of i-AUROC (average i-AUROC) and i-AP (average i-AP), respectively, across all exemplars in the data. 

The example in Figure \ref{fig:attn_quality} shows that i-AUROC and i-AP accurately reflect the performance when detecting key instances for an individual exemplar---their values are consistent with the rankings of the attention scores with respect to the ground truth of two key instances. More importantly, in this example, all models correctly predict the exemplar label (=1), but prediction of key instances differs. Together with more results in Section \ref{sec:sigcomp}, this demonstrates that even if a model misidentifies key instances, it will quite possibly make a correct classification, and vice versa. Therefore, classification performance alone is insufficient to assess MI models: it should be complemented by quantitative evaluation of key instance detection performance, whenever possible. 

Note that we omit the models using non-attention-based MIL methods when assessing key instance detection because they are either unable to identify the key instances (\cite{minet18}), or require complicated post-processing to do so (\cite{mil-lstm20}). We additionally omit the MSA benchmark when considering key instance detection, for two reasons. One, there is inconsistency regarding how to properly extract attention scores in the literature relating to the transformer encoder. For instance, \cite{vit21} used ``attention rollout'' whereas \cite{transmil21} used the final MSA-layer to obtain attention scores. It is not yet clear which way should be adopted. Two, as MSA is not a natural pooling method, its extracted attention scores include components that are irrelevant for MI pooling due to the self-attention mechanism. It is also unclear how to better post-process (e.g., re-normalize) the extracted attention scores to properly identify key instances.

\section{Experiments} \label{sec:experiments}

We conducted experiments on the three tasks illustrated in Figure \ref{fig:miv_tasks}, also described in Section \ref{sec:tasks}, based on the publicly available raw data in  (\cite{qmnist19, sigcomp11, fever18}). For each task, we conducted three rounds of experiments and report mean performance and standard errors. Within each round of experiments, by randomly sampling from disjoint subsets of the raw data (i.e., with \emph{non-overlapping} query classes, \emph{cf.}, Section \ref{sec:tasks}), we constructed a different set of train, validation, and test data, respectively. Note that only the test dataset includes the key instances' ID, for evaluating key instance detection performance. In all tables, the best performance is indicated in bold. Appendix~\ref{data_train} has details on the raw data, feature extractors (from \cite{resnet15, effnet21, sbert19, bn21}), and training process used in each task.

For our proposed approach, i.e., CAP, we use ``CAP-VEMA'' and ``CAP-DBA'' to refer to the models with VEMA and ``DBA with $L_1$-distance'' attention functions, respectively. The exception is our study on the QMNIST task, where we evaluated DBA with both $L_1$- and $L_2$-distance, denoted as CAP-DBA-$L_1$ and CAP-DBA-$L_2$ in Tables \ref{tab:res1c} and \ref{tab:res1e}.\footnote{We thank an anonymous reviewer for suggesting the experiments with $L_2$-distance.}

\subsection{Results for QMNIST handwriting verification} \label{sec:qmnist}

We first consider performance of classification and key instance detection using datasets with a fixed training sample size and without controlling bag sizes, and then go on to consider the effect of varying training sample size and (controlled) average bag sizes.

\subsubsection{Performance of classification and key instance detection} \label{sec:qmnist_perf}

\begin{table}[p]
\centering
\caption{QMNIST: classification performance (mean and standard errors), including non-attention-based benchmarks (MI-Net, Bi-LSTM). The best performance is shown in bold.}
\begin{tabular}{@{}l@{\hskip6pt}c@{\hskip3pt}c@{\hskip3pt}c@{\hskip3pt}c@{\hskip3pt}c@{}}
\toprule				
&AUROC&Accuracy&Precision&Recall&F1-score\\ \cline{2-6}
Baseline&0.708$\pm$0.006&0.658$\pm$0.003&0.647$\pm$0.003&0.637$\pm$0.003&0.637$\pm$0.003\\
GABMIL
&0.660$\pm$0.017&0.623$\pm$0.011&0.613$\pm$0.009&0.613$\pm$0.009&0.613$\pm$0.009\\
PMA
&0.641$\pm$0.005&0.604$\pm$0.010&0.600$\pm$0.006&0.600$\pm$0.006&0.597$\pm$0.009\\
MSA
&0.634$\pm$0.008&0.596$\pm$0.004&0.593$\pm$0.007&0.597$\pm$0.003&0.590$\pm$0.006\\ \midrule
MI-Net
&	0.637$\pm$0.006	&	0.593$\pm$0.005	&	0.597$\pm$0.003	&	0.597$\pm$0.003	&	0.593$\pm$0.003	\\ 
Bi-LSTM
&	0.630$\pm$0.005	&	0.594$\pm$0.010	&	0.590$\pm$0.006	&	0.590$\pm$0.006	&	0.587$\pm$0.009	\\ \midrule
CAP-VEMA(ours)&0.736$\pm$0.004&0.675$\pm$0.007&0.663$\pm$0.007&0.667$\pm$0.009&0.667$\pm$0.009\\
CAP-DBA-$L_1$(ours) & 0.731$\pm$0.011 & 0.675$\pm$0.009 & 0.667$\pm$0.007 & 0.667$\pm$0.007 & 0.667$\pm$0.007\\
CAP-DBA-$L_2$(ours)&\textbf{0.742$\pm$0.004}&\textbf{0.680$\pm$0.004}&\textbf{0.677$\pm$0.003}&\textbf{0.673$\pm$0.003}&\textbf{0.673$\pm$0.003}\\
\bottomrule 
\end{tabular}
\label{tab:res1c}
\end{table}

\begin{table}[p]
\center
     \caption{Classification performance of CAP-VEMA model on test QMNIST dataset broken down by digits}
     \begin{threeparttable}
    \begin{tabular}{cccccm{2cm}}
    \toprule
        Digit&  TP\tnote{1}&  TN\tnote{1}&  FP\tnote{1} &  FN\tnote{1} & Accuracy (=TP+TN) \\ \midrule
        ``0''&  0.246 & 0.439 & 0.148 & 0.167 & \multicolumn{1}{c}{0.685} \\
        ``1''&  0.282 & 0.369 & 0.191 & 0.158 & \multicolumn{1}{c}{0.651} \\
        ``2''&  0.266 & 0.451 & 0.137 & 0.146 & \multicolumn{1}{c}{0.717} \\
        ``3''& 0.231 & 0.445 & 0.142 & 0.182 & \multicolumn{1}{c}{0.676} \\
        ``4''& 0.259 & 0.421 & 0.162 & 0.158 & \multicolumn{1}{c}{0.681} \\
        ``5''& 0.261 & 0.422 & 0.161 & 0.156 & \multicolumn{1}{c}{0.683} \\
        ``6''& 0.318 & 0.395 & 0.184 & 0.103 & \multicolumn{1}{c}{0.713} \\
        ``7''& 0.292 & 0.412 & 0.158 & 0.138 & \multicolumn{1}{c}{0.704} \\
        ``8''& 0.238 & 0.417 & 0.170 & 0.175 & \multicolumn{1}{c}{0.655} \\
        ``9''& 0.273 & 0.385 & 0.190 & 0.152 & \multicolumn{1}{c}{0.658} \\ \bottomrule
    \end{tabular}
    \begin{tablenotes}
    \item[1] TP: True Positive; TN: True Negative; FP: False Positive; FN: False Negative. All figures are relative rate over total number of samples for a specific digit.
    \end{tablenotes}
    \end{threeparttable}
    \label{tab:res1d}
\end{table}

\begin{table}[p]
\center
    \caption{QMNIST: key instance detection performance (mean and standard errors). The best performance is shown in bold.}\label{tab:res1e}
    \begin{tabular}{@{}lccc@{}} 
        \toprule
        	&&	Avg. i-AUROC			&	Avg. i-AP		\\ 
        \cline{3-4}									
        Baseline	&&	0.696	$\pm$	0.001	&	0.619	$\pm$	0.003	\\ 
        GABMIL	&&	0.506	$\pm$	0.001	&	0.479	$\pm$	0.002	\\ 
        PMA	&&	0.509	$\pm$	0.004	&	0.487	$\pm$	0.004	\\
         \midrule									
        CAP-VEMA && 0.832 $\pm$ 0.005 &\textbf{0.784 $\pm$ 0.007} \\ 
        CAP-DBA-$L_1$ &&	0.825	$\pm$	0.006	&	0.771	$\pm$	0.005	\\
        CAP-DBA-$L_2$ && \textbf{0.835 $\pm$ 0.002} & \textbf{0.784 $\pm$ 0.003}\\
        \bottomrule 
    \end{tabular}
 \end{table}

Table \ref{tab:res1c} reports the classification performance of all methods, including the two non-attention-based ones (MI-Net and Bi-LSTM), on QMNIST using AUROC, accuracy, precision, recall, and F1-score.
\footnote{We use the macro average for precision, recall, F1-score because either class may be of interest, e.g., the exemplar label 0 when models are used for anomaly detection, and the label 1 for verification purposes.} It shows the baseline outperforms the benchmark models, while the CAP-based models improve on all measures by substantial margins when compared with the baseline and benchmarks. Interestingly, the results using $L_2$-distance, at least on the QMNIST datasets, are slightly better than those using DBA-$L_1$ and VEMA.

To better understand the exemplar-level verification performance in Table \ref{tab:res1c}, it is worth noting that even though MNIST is generally regarded as a simple dataset, the MI verification (MIV) task based on it can be highly challenging. To put it in context, for the well-known task of image classification to recognize MNIST digits, usually deemed a ``solved'' problem in the literature, a ResNet-18 model can typically achieve above 95\% classification accuracy (\cite{resnet18mnist23}). In contrast, the accuracy of all models in Table \ref{tab:res1c}, using ResNet-18 as a feature extractor, is below 70\% on the MIV task. In addition, our experimental design, constructing each exemplar using the same digit, enables us to break down performance by digits, as shown in Table \ref{tab:res1d}, which facilitates understanding and disentangling relevant factors. Table \ref{tab:res1d} has performance per digit evaluated using the CAP-VEMA model based on one representative test dataset of this task. 

We offer some intuition about why the MNIST MIV task can be difficult and draw some observations from Table \ref{tab:res1d}. Intuitively, the high achievable accuracy, and thus easiness, of the digit recognition task demonstrates that the writing styles of different writers are generally consistent---otherwise inconsistent or ambiguous writing patterns would have made it much more difficult to recognize or differentiate between digits (\cite{dirtymnist21, ambmnist23}). Nonetheless, such consistency of writing styles is in fact a major source of difficulty for a verification task, because the goal of verification is to tell apart different writers' handwriting based on variations in writing style.\footnote{In an extreme case, had all writers' writing style been identical, it would have made the digit recognition task trivial, but made the verification task impossible---that is, close to random guess---even for humans.} MIV adds further challenge by introducing bags of different writers' handwriting. This exacerbates the issue of less variation in writing styles since only one confounding writer in a bag can result in the signal being overwhelmed by the noise. Table \ref{tab:res1d} contains some supporting evidence for this intuition: the digits usually written in a uniform pattern, like ``1'' and ``8'', are more difficult for MIV, leading to higher false positive/negative rates and thus worse accuracy. Conversely, digits that can be naturally written in diverse ways, such as ``2'' and ``7'', can be more accurately verified.

To enable evaluation of key instance detection performance, Table \ref{tab:res1e} shows average i-AUROC and i-AP for the baseline, two benchmarks, and our models. CAP-based models outperform the other three by even larger margins than in the case of classification performance, indicating a superior ability to identify key instances based on the attention scores. Notably, the two benchmarks produce significantly worse performance than the baseline, supporting the hypothesis that failure of their attention mechanism to incorporate the query diminishes their ability to effectively tackle verification tasks of this type.

\subsubsection{Effects of training sample size and bag size} \label{sec:qmnist_ss}

\begin{figure}[t!]
     \centering
     \textbf{10 instances per target bag on average}\\
     \begin{subfigure}[c]{.45\textwidth}
         \centering
         \includegraphics[width=\textwidth, left]{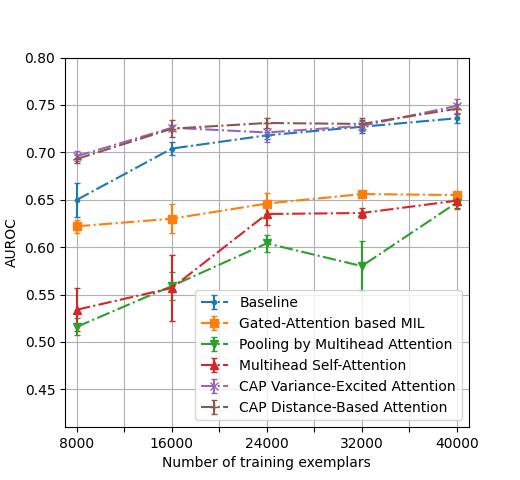} 
         \caption{Classification performance: AUROC}
         \label{qmnistss_cls}
     \end{subfigure}
     \begin{subfigure}[c]{.45\textwidth}
         \includegraphics[width=\textwidth, left]{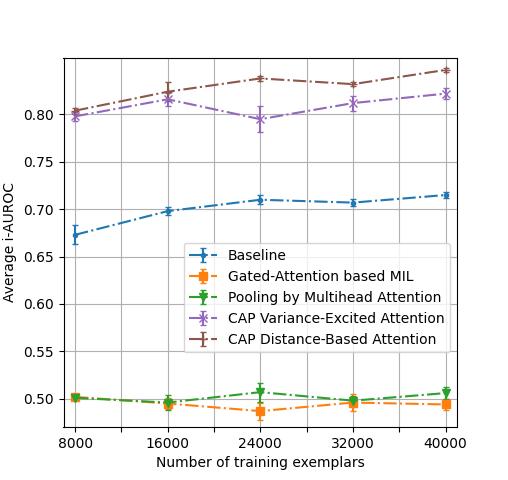} 
         \caption{Key instance detection performance: average i-AUROC}
         \label{qmnistss_exp}
     \end{subfigure}
     \begin{subfigure}[c]{0.45\textwidth}
         \centering
         \includegraphics[width=\textwidth]{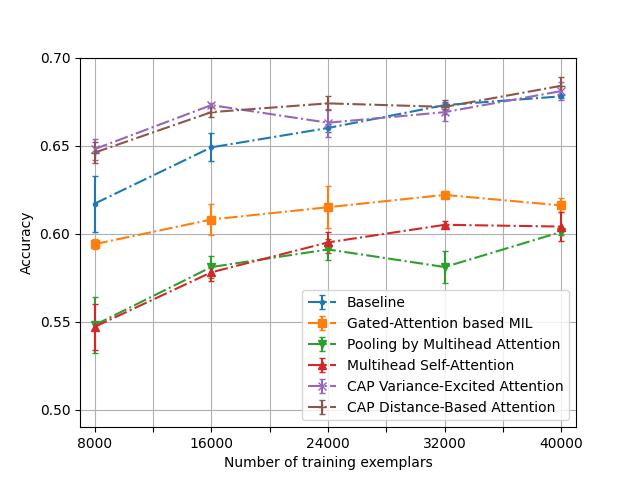}
         \caption{Classification performance: accuracy}
         \label{qmnistss_cls_add}
     \end{subfigure}
     \begin{subfigure}[c]{0.45\textwidth}
         \includegraphics[width=\textwidth]{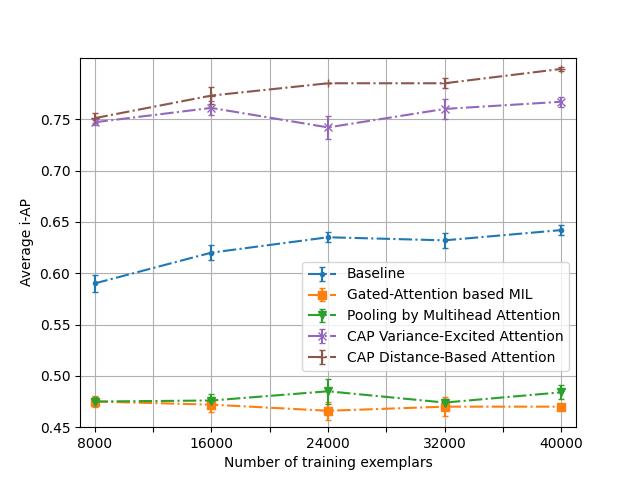}
         \caption{Key instance detection performance: average i-AP}
         \label{qmnistss_exp_add}
     \end{subfigure}
     \caption{Test AUROC (\ref{qmnistss_cls}), accuracy (\ref{qmnistss_cls_add}), average i-AUROC (\ref{qmnistss_exp}), average i-AP (\ref{qmnistss_exp_add}) for QMNIST when varying numbers of training exemplars, from 8,000 to 40,000, with an average of 10 instances per target bag.}
     \label{fig:qmnist_ss}
\end{figure}

Similar to \cite{gabmil18}, we vary the number of exemplars (training sample size) and the number of instances per target bag (bag size) in the training data, and study the effect on model performance. More specifically, we consider mean bag sizes of 10, 20, and 50, with variance 2, 4, and 10, respectively. For mean bag size 10, Figure~\ref{fig:qmnist_ss} shows performance of classification (AUROC, accuracy) and key instance detection (average i-AUOROC, average i-AP), obtained by varying the training sample size from 8,000 to 40,000. The results for mean bag sizes 20 and 50 can be found in Figures \ref{fig:qmnist20ss},~\ref{fig:qmnist50ss} of Appendix~\ref{qmnist_larger_bags} and are very similar. In this experiment, we only consider $L_1$-distance for the DBA attention function.

Based on these results, it is clear that CAP-based models perform substantially better than the other models, for small and large training sample sizes.
The improvement in key instance detection performance (\emph{cf.} Figure \ref{qmnistss_exp} and \ref{qmnistss_exp_add}) is consistent with the hypothesis that this is the main driver for the improvement in classification performance. 
For average target bag sizes of 20 and 50 instances, similar to the case of bag size 10, our models still significantly outperform the other models by substantial margins, according to all criteria. 

\begin{table}[t]
\begin{tabular}{ccccccccc}
\multirow{2}[2]{100pt}{Average number of instances per bag} & & \multicolumn{3}{c}{Number of key instances} & & \multicolumn{3}{c}{$\Big(\frac{\text{Number of key instances}}{\text{Number of instances}}\Big)$} \\[6pt]
\cline{3-5}\cline{7-9} \\[-12pt]
& & Mean & Median & Maximum & & Mean & Median & Maximum \\
\toprule
10 & & 2.742 & 3.0 & 8 & & 0.288 &   0.273 &  0.8 \\
20 & & 3.699 & 4.0 & 15 & & 0.187 &  0.182 & 0.682 \\
50 & & 3.273 & 3.0 & 10 & & 0.066 &  0.061 & 0.222 \\
\bottomrule 
\end{tabular}
\caption{Statistics of ``Number of key instances'' and ``Ratio of (Number of key instances / Number of instances)'' with, on average, 10, 20, and 50 instances per target bag.}
\label{tab:qmnist_ss}
\end{table}

We also observe that all models' performance worsens when the bag size increases. A deeper inspection of the data reveals the reason. In this task, the source of key instances, i.e., a set of handwritten examples of a specific digit from a writer, is limited in the raw data. Because we draw random samples, without replacement and by conditioning on a digit, to generate a bag, key instances in a bag become relatively scarce when we increase the bags size. This can be clearly seen in the second, third, and fourth column of Table \ref{tab:qmnist_ss}, which present the mean, median, and maximum, respectively, for the number of key instances per target bag in the training data: when the average bag size increases, the number of key instances per bag remains stable. Consequently, enlarging the bag size entails adding more non-key instances, or ``noise'', to the bag. In other words, the ratio  ``Number of key instances / Total number of instances'' per bag,\footnote{This ratio stands for the proportion of key instances in each bag, sometimes also known as ``witness rate'' (\cite{milsurvey18, milinterpret22}).} representing a form of signal-to-noise ratio, decreases with increased bag size, see the rightmost three columns of Table \ref{tab:qmnist_ss}. Therefore, in this task, the bag size effectively becomes a proxy for the signal-to-noise ratio. When complicated by low inter-class variation relative to intra-class variation in QMNIST, illustrated in Section~\ref{sec:qmnist_perf}, a low signal-to-noise ratio in the training data creates a more difficult verification task for \emph{any} model or even humans, not just for the models considered here. Empirically, comparing Figures~\ref{fig:qmnist20ss} and \ref{fig:qmnist50ss} to Figure~\ref{fig:qmnist_ss}, the greater difficulty due to larger bag size manifests itself in a drop in performance across \emph{all} models. Similar observations, that a lower signal-to-noise ratio (i.e., witness rate) correlates with increased difficulty of tasks and weaker performance of \emph{all} models, can also be found in the existing MIL literature (e.g.,~\cite{milinterpret22}).

Importantly, we note that the performance of the CAP-based models tends to be \emph{less sensitive} to increases in bag size, indicating that these models are more resilient to lower signal-to-noise ratios. For example, when the average bag size increases from 10 to 20 (\emph{cf.} Figure~\ref{fig:qmnist20ss}), these models' performance varies little while that of the other models  deteriorates substantially. For an average bag size of 50 (\emph{cf.} Figure~\ref{fig:qmnist50ss}), performance of our models decreases only slightly or moderately, while that of other models is noticeably reduced.

\subsection{Results for signature verification against multiple anchors} \label{sec:sigcomp}



\begin{table}[t]
\centering
\caption{Signature verification against multiple anchors: classification performance (mean and standard errors), including two non-attention-based benchmarks (MI-Net and Bi-LSTM). The best performance is shown in bold.}
\begin{tabular}{@{}l@{\hskip6pt}c@{\hskip3pt}c@{\hskip3pt}c@{\hskip3pt}c@{\hskip3pt}c@{}} 
\toprule				
	&	AUROC			&	Accuracy			&	Precision			&	Recall			&	F1-score			\\	
 \cline{2-6}		
Baseline & 0.685$\pm$0.014	&	0.594$\pm$0.008	&	0.697$\pm$0.019	&	0.593$\pm$0.009	&	0.533$\pm$0.015	\\ 
GABMIL
&	0.688$\pm$0.035	&	0.547$\pm$0.018	&	0.697$\pm$0.007	&	0.547$\pm$0.018	&	0.440$\pm$0.036	\\ 
PMA
&\textit{0.756$\pm$0.031}	&	\textit{0.641$\pm$0.025}	&	0.687$\pm$0.033	&	0.643$\pm$0.023	&	0.617$\pm$0.035 \\ 
MSA
&	0.690$\pm$0.011	&	0.619$\pm$0.002	&	0.633$\pm$0.003	&	0.620$\pm$0.000	&	0.610$\pm$0.000	\\ \midrule
MI-Net
&	0.703$\pm$0.039	&	0.643$\pm$0.016	&	0.673$\pm$0.007	&	0.643$\pm$0.017	&	0.623$\pm$0.022	\\ 
Bi-LSTM
&	0.733$\pm$0.027	&	0.653$\pm$0.019	&	0.663$\pm$0.024	&	0.653$\pm$0.020	&	0.647$\pm$0.018	\\ \midrule							
CAP-VEMA(ours)	&\textbf{0.819$\pm$0.006}	&	0.673$\pm$0.027	&	\textbf{0.733$\pm$0.003}	&	\textbf{0.673$\pm$0.029}	&	0.647$\pm$0.041	\\ 
CAP-DBA(ours)	& 0.814$\pm$0.006	&	\textbf{0.674$\pm$0.018}	&	\textbf{0.733$\pm$0.009}	&	\textbf{0.673$\pm$0.020}	&	\textbf{0.653$\pm$0.029}	\\ 
\bottomrule 
\end{tabular}
\label{tab:res2c}
\end{table}

\begin{table}
\center
\caption{Signature verification against multiple anchors: key instance detection performance (mean and standard errors). The best performance is shown in bold.}\label{tab:res2e}
\begin{tabular}{@{}lcc@{}}  
\toprule
	&	Avg. i-AUROC			&	Avg. i-AP			\\ 
\cline{2-3}																	
Baseline	&	0.738	$\pm$	0.021	&	0.678	$\pm$	0.027	\\ 
GABMIL	&	0.533	$\pm$	0.014	&	0.504	$\pm$	0.006	\\ 
PMA	&	0.558	$\pm$	0.011	&	0.517	$\pm$	0.006	\\
 \midrule									
CAP-VEMA	& \textbf{0.995	$\pm$	0.002}	&	\textbf{0.993	$\pm$	0.003}	\\ 
CAP-DBA	&	0.993	$\pm$	0.004	&	0.991	$\pm$	0.005	\\ 
\bottomrule 
\end{tabular}
\end{table}

\begin{figure}[t]
    \centering
    \begin{subfigure}{0.8\textwidth}
        \includegraphics[width=\textwidth]{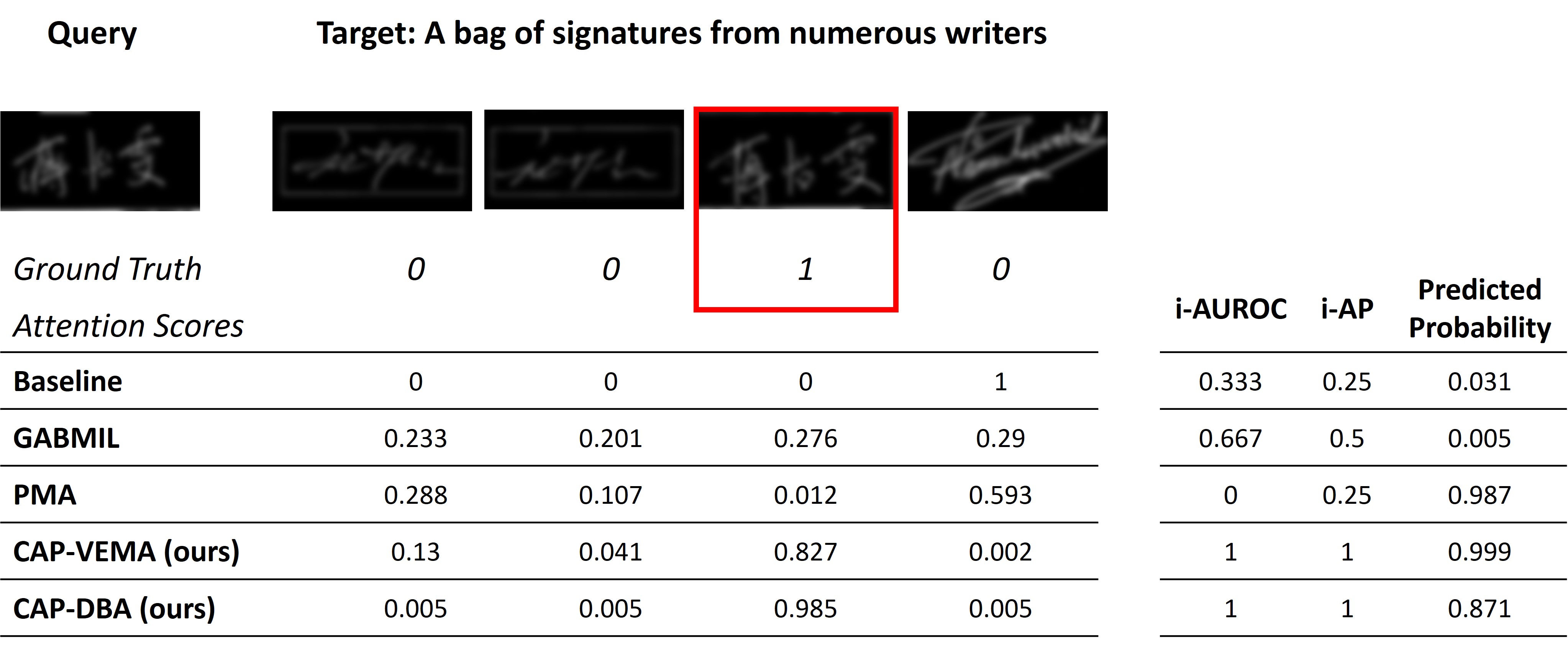}
        \caption{}
        \label{fig:a1_exp1}
        \vspace*{10pt}
    \end{subfigure}
    \begin{subfigure}{0.8\textwidth}
        \includegraphics[width=\textwidth]{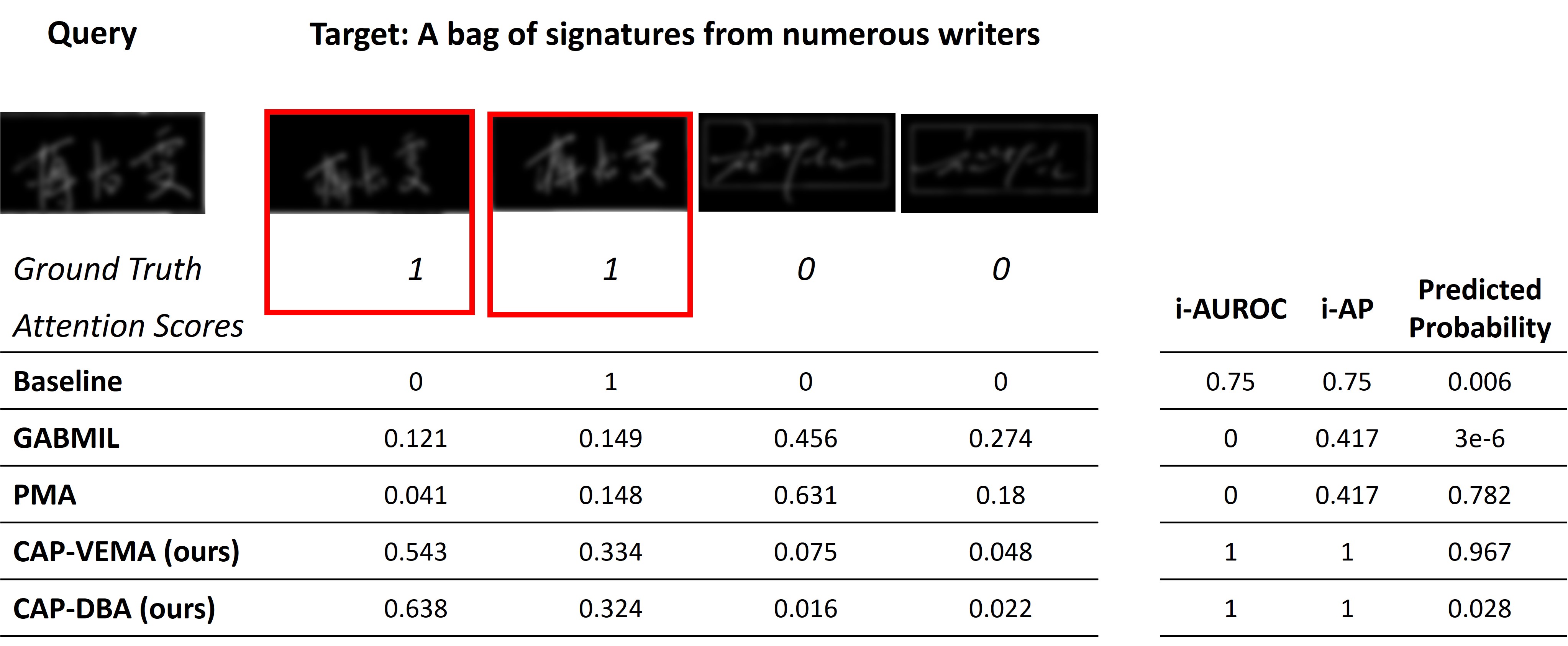}
        \caption{}
        \label{fig:a1_exp2}
    \end{subfigure}
    \caption{More examples to illustrate quantitative evaluations of key instance detection vs. classification. The column ``Predicted Probability'' shows a model's predicted probability of the exemplar label. If it is $\geq 0.5$, this exemplar is classified as 1, otherwise 0. A ground-truth key instance is highlighted by a red rectangle. Signature images are blurred for data use compliance.}
\end{figure}

The main results for signature verification are in Tables \ref{tab:res2c} and \ref{tab:res2e}, respectively. The CAP-based models consistently exceed the performance of the baseline and benchmarks by sizable margins, according to performance of both classification and key instance detection. In particular, they provide near-perfect key instance detection, which supports the hypothesis that their better classification performance is due to their superior ability to correctly identify key instances.

Overall, the pattern of performance in Tables \ref{tab:res2c} and \ref{tab:res2e} is similar to that in the first task, with two exceptions. One, our models' key instance detection performance is close to perfect whilst their classification accuracy is substantially below 100\%. Two, the PMA benchmark appears to perform notably better than the baseline, although with higher standard errors.  Considering the first point, as explained in Section \ref{sec:tasks},  there are two types of negative exemplars in this task: one has an irrelevant query that is completely different from any anchor; another includes a query that is a professional forgery similar to an (authentic) anchor, sometimes looking even more like the anchors than some truly authentic queries. Because the training data do not disclose what type of query yields a negative exemplar (``irrelevant'' or ``forgery'')---to mimic the real-world scenarios where the reasons for rejecting a signature are unavailable---solving this task requires essentially a two-step verification process: (1) to identify the correct anchors from a target bag, if they exist; and (2) to verify whether the query is authentic or a forgery when compared to relevant anchors. This process is the same even if humans perform the verification---the first step may be relatively easy while the second can be highly difficult. The near-perfect key instance detection performance demonstrates our methods' ability to perform the first step well while the relatively low accuracy reflects the difficulty of distinguishing between authentic signatures and forgeries in the second step. 

Considering PMA's relatively high classification accuracy, inspection of its key instance detection performance reveals that ``correct'' classifications are often obtained for the wrong reasons---its key instance detection performance is \emph{far worse} than that of the baseline. This implies that even though some classifications may be ``correct'', the \emph{incorrect} key instances in the target bag are chosen when making those classifications. This faulty behaviour is enabled by the fact that PMA uses a trainable seed vector, rather than the query's feature vector, for MI pooling of the target bag. Intuitively, using incorrect reasoning to obtain classifications implies high variance (standard errors) in classification performance and this is confirmed in Table \ref{tab:res2c}. The discrepancy between performance of classification and key instance detection, highlights the importance of assessing both criteria together.


We show some concrete examples of inconsistency in the accuracy of classifications and key instance detection in Figures \ref{fig:a1_exp1} and \ref{fig:a1_exp2}. In Figure \ref{fig:a1_exp1}, there is one key instance: the third instance. Our models correctly identify this key instance with the highest attention score (which is much higher than the second-highest score) and also make the correct classification. The highest attention score of the baseline and GABMIL benchmark does not identify the correct key instance, although GABMIL's second-highest score does. Note that both i-AUROC and i-AP correctly reflect their worse ability to detect the key instance compared to our models, and as expected, they make an incorrect classification. The PMA benchmark has the poorest key instance detection performance, because its \emph{lowest} attention score corresponds to the actual key instance. Surprisingly, it makes the correct classification with high probability.

For the exemplar in Figure \ref{fig:a1_exp2}, there are two key instances. Our models correctly identify them with the two highest attention scores. Nonetheless, the DBA method makes an incorrect classification, despite the accurate detection of the key instances. Likewise, the baseline correctly identifies one key instance yet still makes an incorrect classification. Both GABMIL and PMA are unable to identify the correct key instances, with the two \emph{lowest} attention scores corresponding to the true key instances. Interestingly, GABMIL makes an incorrect classification, but PMA still makes the correct classification.

\subsection{Results for fact extraction and verification}\label{sec:fever}

\begin{table}[h]
\centering
\caption{Fact extraction and verification (FEVER): classification performance (mean and standard errors), including two non-attention-based benchmarks (MI-Net and Bi-LSTM). The best performance is shown in bold.}
\begin{tabular}{@{}l@{\hskip6pt}c@{\hskip3pt}c@{\hskip3pt}c@{\hskip3pt}c@{\hskip3pt}c@{}} 
\toprule				
&AUROC&Accuracy&Precision&Recall&F1-score\\
\cline{2-6}	
Baseline&0.864$\pm$0.004&0.778$\pm$0.003&0.780$\pm$0.000&0.777$\pm$0.003&0.777$\pm$0.003\\ 
GABMIL
&0.864$\pm$0.003&0.777$\pm$0.002&0.780$\pm$0.000&0.777$\pm$0.003&0.777$\pm$0.003\\ 
PMA
&0.855$\pm$0.002&0.770$\pm$0.002&0.773$\pm$0.003&0.770$\pm$0.000&0.770$\pm$0.000\\ 
MSA
&0.857$\pm$0.001&0.771$\pm$0.003&0.767$\pm$0.003&0.767$\pm$0.003&0.767$\pm$0.003\\ \midrule
MI-Net
&	0.857$\pm$0.002	&	0.769$\pm$0.002	& 0.770$\pm$0.000	&	0.770$\pm$0.000	&	0.770$\pm$0.000 \\ 
Bi-LSTM
&	0.860$\pm$0.002	&	0.771$\pm$0.001	&	0.770$\pm$0.000	&	0.770$\pm$0.000	&	0.770$\pm$0.000	\\ \midrule
CAP-VEMA(ours)&0.886$\pm$0.002&0.807$\pm$0.003&0.813$\pm$0.003&0.807$\pm$0.003&0.807$\pm$0.003\\ 
CAP-DBA(ours)& \textbf{0.898$\pm$0.001} & \textbf{0.817$\pm$0.003} & \textbf{0.827$\pm$0.003} & \textbf{0.817$\pm$0.003} & \textbf{0.817$\pm$0.003} \\ 
\bottomrule 
\end{tabular}
\label{tab:res3c}
\end{table}

\begin{table}
\center
\caption{Fact extraction and verification (FEVER): key instance detection performance (mean and standard errors). The best performance is shown in bold.}\label{tab:res3e}
\begin{tabular}{@{}lcc@{}} 
 \toprule
	&	Avg. i-AUROC			&	Avg. i-AP			\\ 
\cline{2-3}	
Baseline	&	0.584	$\pm$	0.035	&	0.384	$\pm$	0.050	\\ 
GABMIL	&	0.644	$\pm$	0.013	&	0.461	$\pm$	0.017	\\ 
PMA	&	0.620	$\pm$	0.012	&	0.444	$\pm$	0.009	\\
 \midrule									
CAP-VEMA	&	0.835	$\pm$	0.001	&	0.719	$\pm$	0.002	\\ 
CAP-DBA	& \textbf{0.851	$\pm$	0.004}	&	\textbf{0.743	$\pm$	0.005}	\\ 
\bottomrule 
\end{tabular}
\end{table}

Performance on the FEVER task is shown in Tables \ref{tab:res3c} and \ref{tab:res3e}. As in other tasks, CAP-based models outperform the baseline and benchmarks by sizable margins---particularly large margins in key instance detection performance. Interestingly, DBA provides slight but consistent improvements over VEMA across all measures. Similarly to the second task, there is some inconsistency between the quality of classifications and key instance detection when comparing the benchmarks with the baseline.

\subsection{Ablation study} \label{sec:ablation}

\begin{table}[h!]
\begin{subtable}[h]{\textwidth}
\begin{tabular}{lccccc}
\multirow{2}{*}{Model Components} & \multicolumn{2}{c}{Classification} & & \multicolumn{2}{c}{Key instance detection} \\
\cline{2-3}\cline{5-6}
&	AUROC	&	Accuracy	&&	Avg. i-AUROC	&	Avg. i-AP \\
\toprule
\textit{Baseline}	&	\textit{0.708}	&	\textit{0.658}	&&	\textit{0.696}	&	\textit{0.619}	\\
+ New functional form of attention	&	+0.025	&	+0.016	&&	+0.135	&	+0.160	\\
+ Multi-head linear projection	&	+0.000	&	+0.004	&&	-0.003	&	-0.005	\\
+ Co-excitation	&	+0.011	&	+0.007	&&	-0.019	&	-0.017	\\
+ Pre-aggregation LayerNorm	&	-0.008	&	-0.009	&&	+0.023	&	+0.028	\\
\textit{CAP-VEMA}	&	\textit{0.736}	&	\textit{0.675}	&&	\textit{0.832}	&	\textit{0.784}	\\
\bottomrule 
\end{tabular}
\caption{Building CAP-VEMA from the baseline.}
\label{tab:abl_vema_qmnist}
\end{subtable}
\begin{subtable}[h]{\textwidth}
\begin{tabular}{lccccc}
\multirow{2}{*}{Model Components} & \multicolumn{2}{c}{Classification} & & \multicolumn{2}{c}{Key instance detection} \\
\cline{2-3}\cline{5-6}
&	AUROC	&	Accuracy	&&	Avg. i-AUROC	&	Avg. i-AP \\
\toprule
\textit{Baseline}	&	\textit{0.708}	&	\textit{0.658}	&&	\textit{0.696}	&	\textit{0.619}	\\
+ New functional form of attention	&	+0.009	&	+0.005	&&	+0.124	&	+0.147	\\
+ Multi-head linear projection	&	+0.015	&	+0.014	&&	+0.008	&	+0.010	\\
+ Co-excitation	&	+0.004	&	-0.004	&&	+0.004	&	+0.009	\\
+ Pre-aggregation LayerNorm	&	-0.005	&	+0.002	&&	-0.008	&	-0.014	\\
\textit{CAP-DBA}	&	\textit{0.731}	&	\textit{0.675}	&&	\textit{0.825}	&	\textit{0.771}	\\
\bottomrule 
\end{tabular}
\caption{Building CAP-DBA from the baseline.}
\label{tab:abl_dba_qmnist}
\end{subtable}
\caption{Incremental effects of model components by progressively building CAP-VEMA and CAP-DBA from the baseline, in the QMNIST task.}
\label{tab:abl_qmnist}
\end{table}

\begin{table}[t]
\begin{subtable}[h]{\textwidth}
\begin{tabular}{lccccc}
\multirow{2}{*}{Model Components} & \multicolumn{2}{c}{Classification} & & \multicolumn{2}{c}{Key instance detection} \\
\cline{2-3}\cline{5-6}
&	AUROC	&	Accuracy	&&	Avg. i-AUROC	&	Avg. i-AP \\
\toprule
\textit{Baseline}	&	\textit{0.685}	&	\textit{0.594}	&&	\textit{0.738}	&	\textit{0.678}	\\
+ New functional form of attention	&	+0.047	&	+0.012	&&	+0.228	&	+0.277	\\
+ Multi-head linear projection	&	+0.036	&	+0.003	&&	+0.018	&	+0.024	\\
+ Co-excitation	&	+0.043	&	+0.053	&&	+0.007	&	+0.009	\\
+ Pre-aggregation LayerNorm &	+0.008	&	+0.011	&&	+0.005	&	+0.006	\\
\textit{CAP-VEMA}	&	\textit{0.819}	&	\textit{0.673}	&&	\textit{0.995}	&	\textit{0.993} \\
\bottomrule 
\end{tabular}
\caption{Building CAP-VEMA from the baseline.}
\label{tab:abl_vema}
\end{subtable}
\begin{subtable}[h]{\textwidth}
\begin{tabular}{lccccc}
\multirow{2}{*}{Model Components} & \multicolumn{2}{c}{Classification} & & \multicolumn{2}{c}{Key instance detection} \\
\cline{2-3}\cline{5-6}
&	AUROC	&	Accuracy	&&	Avg. i-AUROC	&	Avg. i-AP \\
\toprule
\textit{Baseline}	&	\textit{0.685}	&	\textit{0.594}	&&	\textit{0.738}	&	\textit{0.678}	\\
+ New functional form of attention	&	+0.064	&	+0.013	&&	+0.243	&	+0.299	\\
+ Multi-head linear projection	&	-0.023	&	+0.011	&&	+0.005	&	+0.006	\\
+ Co-excitation	&	+0.073	&	+0.045	&&	+0.006	&	+0.007	\\
+ Pre-aggregation LayerNorm &	+0.015	&	+0.011	&&	+0.001	&	+0.001	\\
\textit{CAP-DBA}	&	\textit{0.814}	&	\textit{0.674}	&&	\textit{0.993}	&	\textit{0.991}	\\
\bottomrule 
\end{tabular}
\caption{Building CAP-DBA from the baseline.}
\label{tab:abl_dba}
\end{subtable}
\caption{Incremental effects of model components by progressively building CAP-VEMA and CAP-DBA from the baseline, in the signature verification task.}
\label{tab:abl_sig}
\end{table}

\begin{table}[ht!]
\begin{subtable}[h]{\textwidth}
\begin{tabular}{lccccc}
\multirow{2}{*}{Model Components} & \multicolumn{2}{c}{Classification} & & \multicolumn{2}{c}{Key instance detection} \\
\cline{2-3}\cline{5-6}
&	AUROC	&	Accuracy	&&	Avg. i-AUROC	&	Avg. i-AP \\
\toprule
\textit{Baseline}	&	\textit{0.864}	&	\textit{0.778}	&&	\textit{0.584}	&	\textit{0.384}	\\
+ New functional form of attention	&	+0.011	&	+0.017	&&	+0.185	&	+0.254	\\
+ Multi-head linear projection	&	+0.003	&	+0.004	&&	+0.039	&	+0.040	\\
+ Co-excitation	&	+0.010	&	+0.008	&&	+0.008	&	+0.012	\\
+ Pre-aggregation LayerNorm	&	-0.002	&	+0.000	&&	+0.020	&	+0.029	\\
\textit{CAP-VEMA}	&	\textit{0.886}	&	\textit{0.807}	&&	\textit{0.835}	&	\textit{0.719}	\\
\bottomrule 
\end{tabular}
\caption{Building CAP-VEMA from the baseline.}
\label{tab:abl_vema_fever}
\end{subtable}
\begin{subtable}[h]{\textwidth}
\begin{tabular}{lccccc}
\multirow{2}{*}{Model Components} & \multicolumn{2}{c}{Classification} & & \multicolumn{2}{c}{Key instance detection} \\
\cline{2-3}\cline{5-6}
&	AUROC	&	Accuracy	&&	Avg. i-AUROC	&	Avg. i-AP \\
\toprule
\textit{Baseline}	&	\textit{0.864}	&	\textit{0.778}	&&	\textit{0.584}	&	\textit{0.384}	\\
+ New functional form of attention	&	+0.018	&	+0.022	&&	+0.206	&	+0.281	\\
+ Multi-head linear projection	&	+0.001	&	+0.005	&&	+0.038	&	+0.046	\\
+ Co-excitation	&	+0.010	&	+0.010	&&	-0.016	&	-0.018	\\
+ Pre-aggregation LayerNorm	&	+0.006	&	+0.003	&&	+0.039	&	+0.050	\\
\textit{CAP-DBA}	&	\textit{0.898}	&	\textit{0.817}	&&	\textit{0.851}	&	\textit{0.743}	\\
\bottomrule 
\end{tabular}
\caption{Building CAP-DBA from the baseline.}
\label{tab:abl_dba_fever}
\end{subtable}
\caption{Incremental effects of model components by progressively building CAP-VEMA and CAP-DBA from the baseline, in the FEVER task.}
\label{tab:abl_fever}
\end{table}

To understand the effects of components from our models, it is instructive to consider the incremental effects of progressively adding CAP components to the baseline, one component at a time, using both VEMA and DBA attention functions. The results of the incremental effects on performance are presented in Tables \ref{tab:abl_qmnist}, \ref{tab:abl_sig}, \ref{tab:abl_fever}, for the three tasks, respectively.

The QMNIST task is synthesized by generating each exemplar using the same digit, making instances within a target bag all look alike. As such, it may be a well-controlled test of the effectiveness of our approach. Table \ref{tab:abl_qmnist} demonstrates that the new functional forms of attention, VEMA and DBA, contribute substantially more to key instance detection performance than the other components, and to a large portion of the improvement in classification performance. Note that in this task, the baseline's performance already significantly exceeds that of the benchmarks (\emph{cf.} Tables \ref{tab:res1c} and \ref{tab:res1e}), possibly because its attention implicitly incorporates information from the query instance. On top of the baseline, VEMA and DBA bring further substantial improvement. This strongly supports the effectiveness of our proposed attention functions in this challenging task. We note that the performance of DBA, despite its simpler form, is comparable to that of VEMA.

For the other two tasks, signature verification and FEVER, the overall patterns are similar: the new functional form of attention is typically the top contributor to model performance. The two components, the attention function and ``co-excitation'', jointly account for the majority of performance, for both classification and key instance detection.

These results demonstrate that most of the increased ability to identify key instances can be attributed to the new functional forms of attention, VEMA and DBA. In conjunction with co-excitation, they also contribute the most to classification enhancement.

On the other hand, the contribution of the multi-head projection is mixed. When moving from the ``non-headed''
version (i.e., without linear projections) to a multi-head version, the performance differences vary across tasks and measures. Additional analyses provided in Appendix~\ref{abl_mh} isolate effects from the multi-head mechanism by comparing models with and without multiple heads. Those analyses demonstrate the limited contribution from the multi-head mechanism, consistent with the results in this section. The reason may be that, in these tasks, each instance is typically an independent and complete object with a holistic---not partial---feature representation, while the multi-head mechanism is designed ``to jointly attend to information from different \emph{representation subspaces} at different positions'' (\cite{transformer17}). Therefore, given the presence of other model components, multi-headedness alone appears to add little value in the tasks considered in our study.\footnote{Note that the transformer used separate multi-head projections for each ``query, key, value'' triplet in attention. Based on the ablation results, we chose not to use separate projections for our experiments. Instead, we opted for less model complexity, by employing a single, shared linear projection in the multi-head formulation of CAP.}

Finally, from Tables~\ref{tab:abl_sig} and \ref{tab:abl_fever} we observe that there are consistent improvements brought by multi-head ``pre-aggregation LayerNorm'' (\emph{cf.} Equations~\eqref{perhead_p}, \eqref{perhead_q}, as the last element in the CAP architecture), compared to no normalization. To check the robustness of CAP to such normalization, we tested the sensitivity of our models to an alternative form of LayerNorm, shown in Appendix~\ref{abl_ln}. This alternative is a standard LayerNorm without multi-headness, akin to the post-attention LayerNorm adopted in the transformer model, applied \emph{after} bag aggregation and \emph{after} all heads are concatenated in Equations~\eqref{perhead_p} and \eqref{perhead_q}. Henceforth, we call it ``post-aggregation LayerNorm''. The results in Appendix~\ref{abl_ln} show similar performance of CAP-based models when the pre-aggregation LayerNorm is replaced with the post-aggregation LayerNorm, and thus demonstrate the robustness of CAP to different forms of output normalization.

\section{Conclusions and future research}
We introduce multi-instance verification (MIV), which combines, but differs from, verification and MIL, and show theoretically and empirically that a lack of information about the query instance in attention is undesirable when pooling the target bag into a bag-level representation. We present a new approach named ``Cross Attention Pooling'' (CAP) that explicitly considers the query in attention, along with two new attention functions within the CAP framework. We also evaluate a simple baseline model and several benchmark models, adapted from SOTA attention-based MIL methods, that omit the query in attention. Results on three different tasks show that the benchmark methods are not better, and sometimes significantly worse, than the simple baseline. In contrast, the CAP-based methods outperform the baseline and benchmarks by sizeable margins in terms of both classification and key instance detection performance. Ablation studies confirm the superior ability of the new attention functions to identify key instances and establish the contributions of the key components of the CAP architecture.

The abstract representation of ``multi-instance verification'', as distinct from traditional MIL, opens up the potential for research on a  broad range of problems and applications. For example, either the ``query'' or the ``target'' may be a single instance, bag of instances, or bag of bags, and they may exhibit different modalities---for example, one may be textual and the other may consist of images. We believe our work provides a generic framework that may inspire research in directions such as new architectures, attention mechanisms, real-world applications, or methods for selecting key instances and evaluating their quality.

As a first step towards tackling MIV, our paper also offers future opportunities connecting to some domain-specific, well-established areas under active research. For example, while this paper considers CV datasets that enable well-controlled experiments, disentangling factors without compromising on the difficulty of tasks, applications of MIV on more complex datasets such as ImageNet or MS-COCO may lead to research in areas of cross-domain few-shot learning, weakly-supervised one-shot object detection, to name a few. For the NLP domain, a comparison of our approach to other methods specifically designed for particular tasks, e.g., FEVER, may also be an interesting topic for prospective research (\emph{cf.} Section \ref{sec:lit_rev}). Beyond CV and NLP, MIV and its solutions may also be applied to potentially broader domains such as time series, multi-modal learning, and so forth. Those generalizations are sufficiently significant and challenging to warrant dedicated, standalone future research, in which domain-specific baselines may be appropriate for comparison, rather than the domain-agnostic MIL models adopted in this paper. Another avenue for future work is to develop methods that scale to large bags or real-time scenarios. Finally, interpretability and its evaluation, in a way that is more general than the ``key instance detection'' performance considered in this paper, is yet another avenue to be explored. Those research topics, typically investigated in the context of domain-specific applications, are left to future work.


\acks
We thank the anonymous reviewers for their helpful comments that enabled us to improve our paper. X. Xu acknowledges financial support from the Machine Learning Group of the Computer Science Department at the University of Waikato. 

\newpage
\appendix

\section{Proof of Proposition \ref{prop1}} \label{prop1proof}
\begin{proof}{[Proposition \ref{prop1}: Variables incorporated in the attention scores]}

Because it is well-known that 
\[H\Big(U\mid V^{Target}\Big) \geq H\Big(U\mid V^{Query}, V^{Target}\Big), \] 
that is, adding more conditioning variables cannot decrease informativeness, to prove Inequality~\eqref{prop1ineq}, we only need to show $H\Big(U\mid V^{Target}\Big) \neq H\Big(U\mid V^{Query}, V^{Target}\Big)$. We prove this by contradiction. 

Based on the well-known property of conditional entropy and mutual information,
\begin{align} \label{mi_prop1}
H(U\mid V^{Query}, V^{Target}) = H(U\mid V^{Target}) - I(U, V^{Query}\mid V^{Target}),
\end{align}
where $I(\cdot,\cdot\mid\cdot)$ denotes the conditional mutual information. It has the well-known non-negativity property, $I\geq 0$, if all random variables are valued in standard Borel spaces.

By symmetry, we can also obtain the following equation analogous to \eqref{mi_prop1}:
\begin{align} \label{mi_prop2}
H(U\mid V^{Query}, V^{Target}) = H(U\mid V^{Query} ) - I(U,V^{Target}\mid V^{Query}).
\end{align}

Subtracting Equation~\eqref{mi_prop2} from Equation~\eqref{mi_prop1} and re-arranging, we get
\begin{align} \label{mi_eq}
H(U\mid V^{Target}) - I(U, V^{Query}\mid V^{Target}) = H(U\mid V^{Query} ) - I(U,V^{Target}\mid V^{Query}).
\end{align}

Now, if $H\Big(U\mid V^{Target}\Big) = H\Big(U\mid V^{Query}, V^{Target}\Big)$, based on Equation~\eqref{mi_prop1}, $I(U, V^{Query}\mid V^{Target}) = 0$, and thus Equation~\eqref{mi_eq} becomes
\begin{align} \label{mi_ineq}
&H(U\mid V^{Target}) = H(U\mid V^{Query} ) - I(U,V^{Target}\mid V^{Query}) \implies \nonumber\\
&H(U\mid V^{Target}) \leq H(U\mid V^{Query})\quad\text{(Non-negativity of $I(U,V^{Target}\mid V^{Query})$)}. 
\end{align}
As a result, Inequality~\eqref{mi_ineq} must always hold, for all values of $V^{Target}, V^{Query}$, implying 
\[Pr\Big(H(U\mid V^{Target}) > H(U\mid V^{Query})\Big)=0,
\] 
which contradicts Assumption \ref{ass1} because the assumption requires
\[Pr\Big(H(U\mid V^{Target}) > H(U\mid V^{Query})\Big)>0.\]
This concludes the proof of Inequality~\eqref{prop1ineq}. By symmetry, Inequality~\eqref{prop2ineq} can also be shown in a similar manner, which completes the proof of Proposition \ref{prop1}.
\end{proof}

\section{Details of data and training process} \label{data_train}

The following details of training are common to all three tasks considered in the experiments.

Prior to full training of all network parameters, we adopt an initial learning stage where we freeze the feature extractor's model weights and train only the parameters specific to the baseline and benchmarks. This initial phase is stopped when the accuracy on the validation dataset does not improve for two epochs.

Subsequently, when all parameters are trained end-to-end, we also adopt early stopping based on validation accuracy, with different stopping criteria for different tasks as discussed below. The optimization for training is conducted using the RMSprop optimizer, with the same parameters for all models: $\rho=0.9, \epsilon=$1e-7. Due to a well-known behavior of ``batch normalization'' (BatchNorm) causing degradation in performance during transfer learning, see, e.g.,~\cite{bn21}, we freeze the BatchNorm layers, if any, in a feature extractor, typically for CV tasks. More precisely, the BatchNorm moving average and the trainable parameters are not updated during training.\footnote{We also tried allowing parameters to be trainable but not updating the moving average parameters and found similar experimental results.}

For each task, we conducted three rounds of experiments for all models considered in our study. All experiments were run on a cluster of four NVIDIA RTX A6000 GPUs, of which the duration varies depending on early stopping triggers. Roughly speaking, a single round of training of one model takes approximately one hour for the QMNIST task, two to three hours for the FEVER task, and three to five hours for the signature verification task.

All models were developed using Tensorflow 2.9.3, with some use of ``TensorFlow official models'' 2.9.0 (\cite{tf_model20}, Apache License 2.0) and scikit-learn 1.2.0 (\cite{sklearn}, BSD License). The original license and terms-of-use of the assets (i.e., raw data, pre-trained feature extractors) used in our research are listed alongside each asset in the following.

\subsection{QMNIST handwriting verification}
We collected QMNIST data (BSD-style license, \url{https://github.com/facebookresearch/qmnist/blob/main/LICENSE}) according to \cite{qmnist19} and from links therein. QMNIST was created to be similar in structure to MNIST, with a training set composed of 60,000 images of digits (``Train/60K'') and a test set of 10,000 digits (``Test/10K''). These two sets turn out to have non-overlapping writer-IDs, which is ideal for our requirement that test writer's handwriting is not seen during training. We constructed a training dataset by randomly selecting from the ``Train/60K'' set, including both the query and target bag. To construct the validation and test datasets, we first split ``Test/10K'' into two sets with non-overlapping writer-IDs and then drew random samples from each set. By construction, the sample sizes of the train/dev/test datasets are 21,509/2,408/2,253 respectively (the same sample sizes for all rounds of experiments were used to draw random samples), and the proportions of their exemplar labels are approximately 50:50. In an exemplar, the bag size $N$ varies between 3 and 25, with a mean of $6.9$ instances per target bag and a variance of $6.4$ approximately.

We employed ResNet-18 (\cite{resnet15}) as the feature extractor, initialized by a set of pre-trained weights based on ImageNet. The pre-trained weights were downloaded from \url{https://pypi.org/project/image-classifiers} (MIT License). Prior to inputting the QMNIST images into ResNet-18, we resized them from their original size of $(28\times28)$ to $(32\times32)$, because this is the minimum input size required by ResNet-18. After putting an image through ResNet-18, we obtained its feature vector by applying a ``Global Average Pooling'' (GAP) on the output of the feature extractor's penultimate layer.

The learning rate of the RMSprop optimizer was piece-wise constant: 1e-4 for the first 5 epochs, 5e-5 for the next 15 epochs, and 2e-5 for the remaining epochs. The mini-batch size was 768, and the early-stopping criterion was non-improvement of validation accuracy for 30 epochs. The number of heads was two for multi-head attention, when applicable.

\subsection{Signature verification against multiple anchors}
We collected raw images of signatures, both authentic and forged, based on \cite{sigcomp11} and from the link \url{http://www.iapr-tc11.org/mediawiki/index.php/ICDAR_2011_Signature_Verification_Competition_(SigComp2011)} (data disclaimer: \url{http://www.iapr-tc11.org/dataset/ICDAR_SignatureVerification/SigComp2011/disclaimer.pdf}). We resized all signature images from their original sizes to $(512\times256)$ and converted them from RGB to black-and-white. To construct the train, validation, and test datasets for each round of the experiment, we first split the raw data into three sets with non-overlapping writer-IDs and then drew random samples from each set. The sample size for training is around 82,000, and for validation/test is close to 10,000. By construction, the proportion of exemplar labels in all datasets is approximately 50:50. In an exemplar, the bag size $N$ varies between two and eight.

For the feature extractor, we employed one from the EfficientNetV2 family (\cite{effnet21}), called ``EfficientNetV2B3'', initialized with ImageNet-pretrained weights. The pretrained weights were downloaded from \url{https://www.tensorflow.org/versions/r2.9/api_docs/python/tf/keras/applications/efficientnet_v2/EfficientNetV2B3} (CC-BY-4.0 license). We resized all the signature images to an identical size of $(256\times128)$. After putting an image through EfficientNetV2B3, we obtained its feature vector by applying global average pooling on the output from the penultimate layer. 

The learning rate of the RMSprop optimizer was piece-wise constant: 5e-6 for the first 5 epochs, 2e-6 for the next 5 epochs, and 1e-6 for the remaining epochs. The mini-batch size was 64, and the early-stopping criterion for training was non-improving validation accuracy for 10 epochs. The number of heads was set to six for multi-head attention.

\subsection{Fact extraction and verification}
For``fact extraction and verification'' (FEVER), we collected the raw data of claims and evidence as in \cite{fever18} and from the FEVER (2018) website \url{https://fever.ai/dataset/fever.html} (license see \url{https://fever.ai/download/fever/license.html}). The FEVER raw data was already split into training, validation, and test data, based on a pre-processed dump (June 2017) of Wikipedia pages---more precisely, the corresponding links are: \url{https://fever.ai/download/fever/train.jsonl}, \url{https://fever.ai/download/fever/paper_dev.jsonl}, \url{https://fever.ai/download/fever/paper_test.jsonl}, \url{https://fever.ai/download/fever/wiki-pages.zip}. For all three rounds of experiments, we used the full set of raw validation/test data (6616/6613 exemplars respectively) as our validation/test datasets. To construct our training dataset in each round of experiments, we randomly sampled 33,000 exemplars from the raw training set, with the proportion of the exemplar labels being approximately 50:50. In an exemplar, we truncated the amount of evidence in a bag, i.e., the bag size $N$, if it was greater than 47, and capped the number of tokens in a piece of evidence to 96.\footnote{The truncations are purely for technical reasons (to limit the use of memory/computing), and we found it impacts less than 10\% of the total samples.} 

For the feature extractor, we employed one from the Sentence-BERT family (SBERT,~\cite{sbert19}), called ``multi-qa-MiniLM-L6-cos-v1'', initialized with weights pretrained based on 215M question-answer pairs from various sources. We also used SBERT's native tokenizer to pre-process all our datasets. See more details, downloadable weights, and the tokenizer at \url{https://www.sbert.net/docs/pretrained_models.html} (Apache License 2.0, \url{https://github.com/UKPLab/sentence-transformers/blob/master/LICENSE}). We obtained the feature vector of any textual paragraph by putting it through the feature extractor, and retrieved the vector of SBERT's special token ``[CLS]''\footnote{Note the vector of SBERT's special token ``[CLS]'' is different from that of the MSA benchmark's ``[CLS]''.} from the models' penultimate layer.

The learning rate of the RMSprop optimizer was piece-wise constant: 1e-5 for the first 10 epochs, 5e-6 for the next 10 epochs, and 2e-6 for the remaining epochs. The mini-batch size was 48, and the early-stopping criterion was non-improving validation accuracy for 10 epochs. The number of heads was set to four for multi-head attention.

\begin{figure}[t]
     \centering
     \textbf{20 instances per target bag on average}\\
     \begin{subfigure}[c]{.45\textwidth}
         \centering
         \includegraphics[width=\textwidth, left]{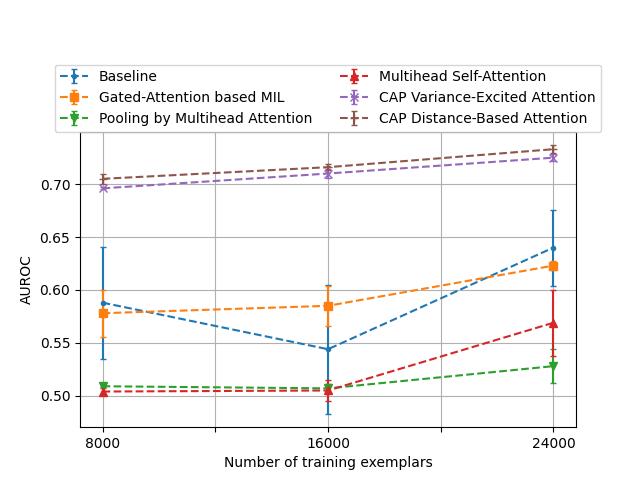} 
         \caption{Classification performance: AUROC}
         \label{qmnistss20cls}
     \end{subfigure}
     \begin{subfigure}[c]{.45\textwidth}
         \includegraphics[width=\textwidth, left]{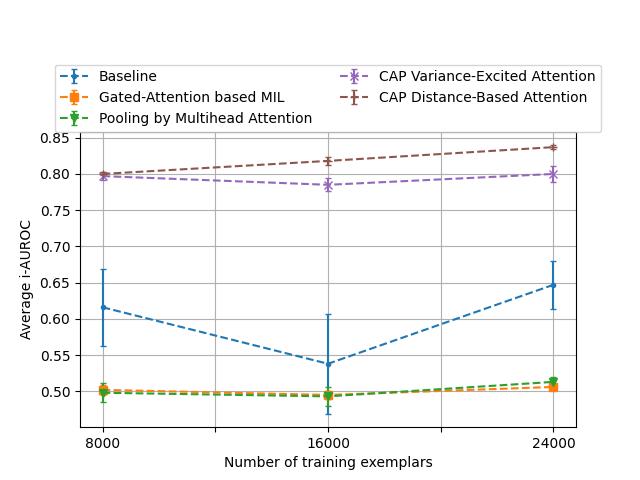} 
         \caption{Key instance detection: average i-AUROC}
         \label{qmnistss20exp}
     \end{subfigure}
     \begin{subfigure}[c]{0.45\textwidth}
         \centering
         \includegraphics[width=\textwidth]{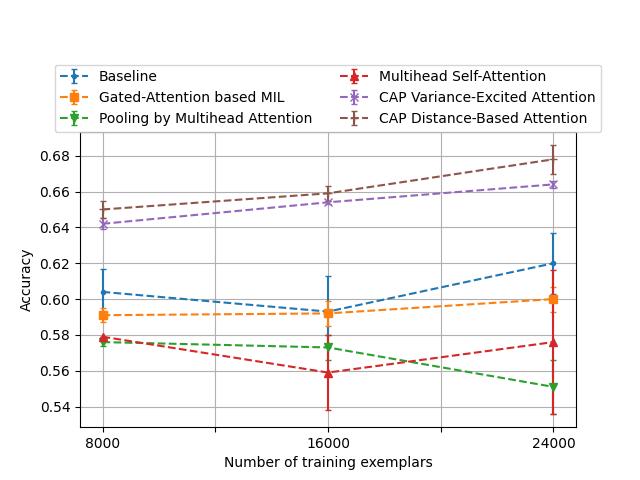}
         \caption{Classification performance: accuracy}
         \label{qmnistss20cls_add}
     \end{subfigure}
     \begin{subfigure}[c]{0.45\textwidth}
         \includegraphics[width=\textwidth]{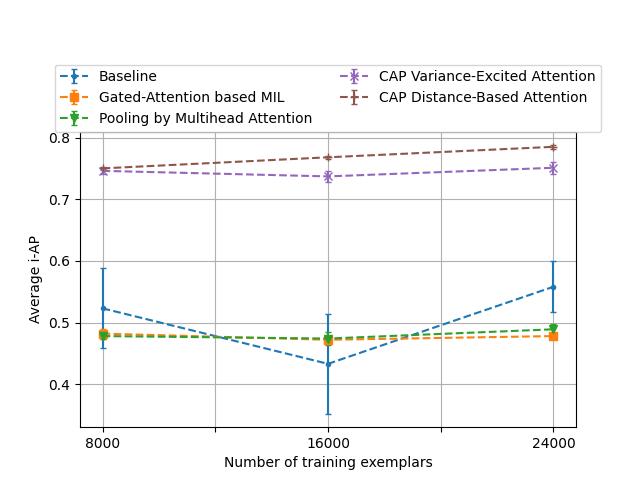}
         \caption{Key instance detection: average i-AP}
         \label{qmnistss20exp_add}
     \end{subfigure}
     \caption{Test AUROC (\ref{qmnistss20cls}), accuracy (\ref{qmnistss20cls_add}), average i-AUROC (\ref{qmnistss20exp}), average i-AP (\ref{qmnistss20exp_add}) for QMNIST for varying numbers of training exemplars, from 8,000 to 24,000, with an average of 20 instances per target bag.}
     \label{fig:qmnist20ss}
\end{figure}

\begin{figure}[h!]
\vspace{-10mm}
     \centering
     \textbf{50 instances per target bag on average}\\
     \begin{subfigure}[c]{.45\textwidth}
         \centering
         \includegraphics[width=\textwidth, left]{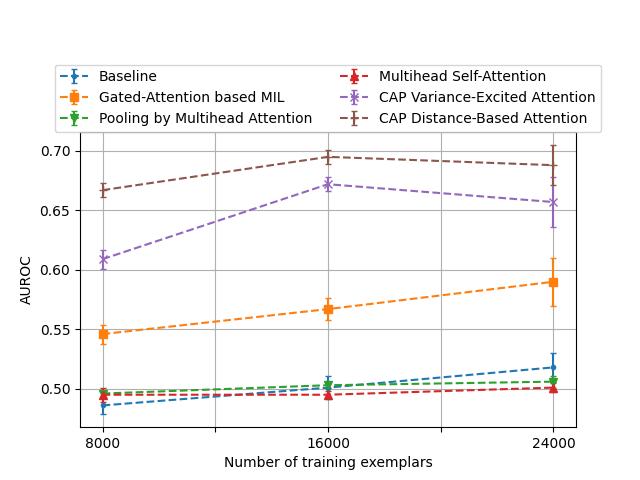} 
         \caption{Classification performance: AUROC}
         \label{qmnistss50cls}
     \end{subfigure}
     \begin{subfigure}[c]{.45\textwidth}
         \includegraphics[width=\textwidth, left]{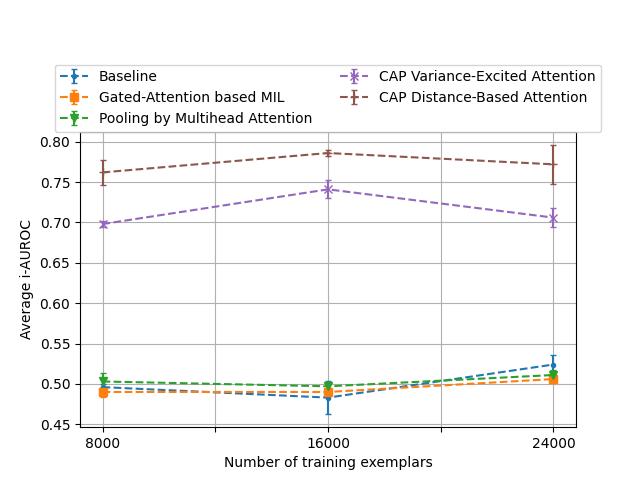} 
         \caption{Key instance detection: average i-AUROC}
         \label{qmnistss50exp}
     \end{subfigure}
     \begin{subfigure}[c]{0.45\textwidth}
         \centering
         \includegraphics[width=\textwidth]{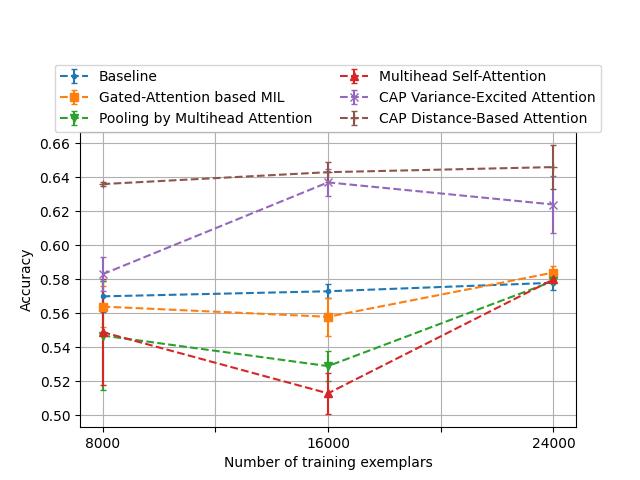}
         \caption{Classification performance: accuracy}
         \label{qmnistss50cls_add}
     \end{subfigure}
     \begin{subfigure}[c]{0.45\textwidth}
         \includegraphics[width=\textwidth]{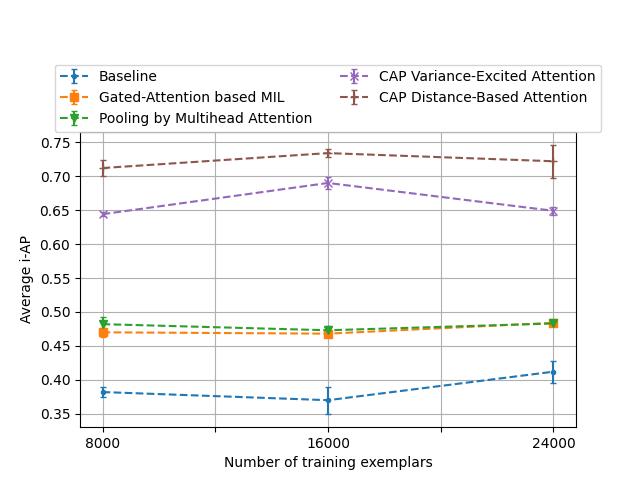}
         \caption{Key instance detection: average i-AP}
         \label{qmnistss50exp_add}
     \end{subfigure}
     \caption{Test AUROC (\ref{qmnistss50cls}), accuracy (\ref{qmnistss50cls_add}), average i-AUROC (\ref{qmnistss50exp}), average i-AP (\ref{qmnistss50exp_add}) for QMNIST for varying numbers of training exemplars, from 8,000 to 24,000, with an average of 50 instances per target bag.}
     \label{fig:qmnist50ss}
\end{figure}

\begin{figure}[h!]
     \centering
     \begin{subfigure}[t]{0.49\textwidth}
         \centering
         \includegraphics[width=\textwidth, left]{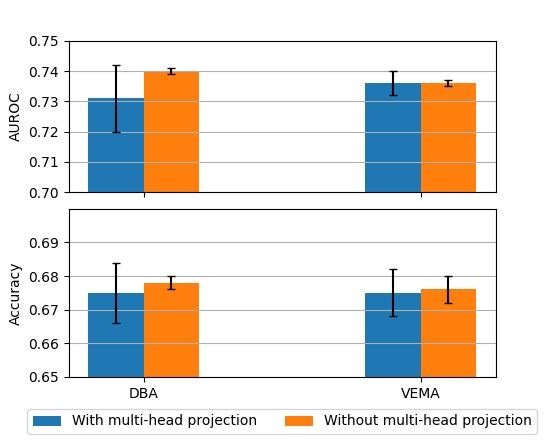}
         \caption{Classification}
         \label{abl_attnmh_cls_qmnist}
     \end{subfigure}
     \hfill
     \begin{subfigure}[t]{0.49\textwidth}
         \includegraphics[width=\textwidth]{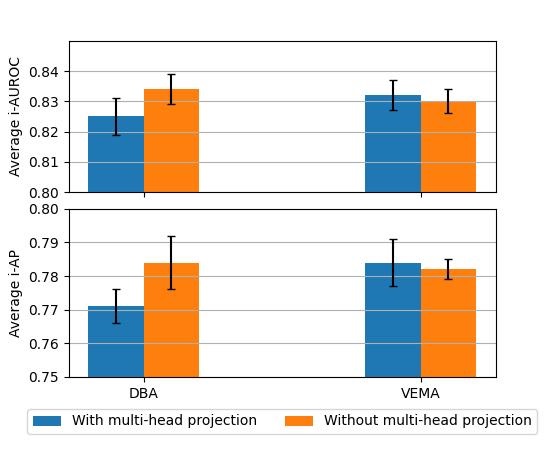}
         \caption{Key instance detection}
         \label{abl_attnmh_exp_qmnist}
     \end{subfigure}
     \caption{Model performance for the QMNIST task under our CAP framework (DBA and VEMA), with and without the multi-head linear projections. Left column (\ref{abl_attnmh_cls_qmnist}) is classification performance; right column (\ref{abl_attnmh_exp_qmnist}) shows performance of detecting key instances.
     }
     \label{fig:abl_attnmh_qmnist}
\vspace{-15mm}
\end{figure}

\section{QMNIST: Results for larger average bag sizes} \label{qmnist_larger_bags}

Learning curves for the QMNIST data for mean bag sizes 20 and 50 are shown in Figures \ref{fig:qmnist20ss} and \ref{fig:qmnist50ss} respectively, in the range of 8,000 to 24,000 training exemplars.

\section{Does multi-head attention matter? } \label{abl_mh}

\begin{figure}[ht!]
     \centering
     \begin{subfigure}[t]{0.49\textwidth}
         \centering
         \includegraphics[width=\textwidth, left]{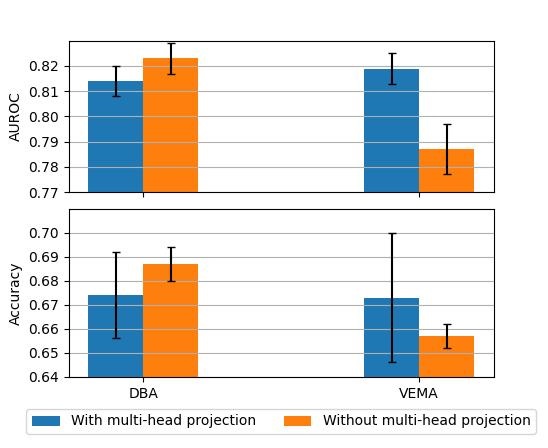}
         \caption{Classification}
         \label{abl_attnmh_cls}
     \end{subfigure}
     \hfill
     \begin{subfigure}[t]{0.49\textwidth}
         \includegraphics[width=\textwidth]{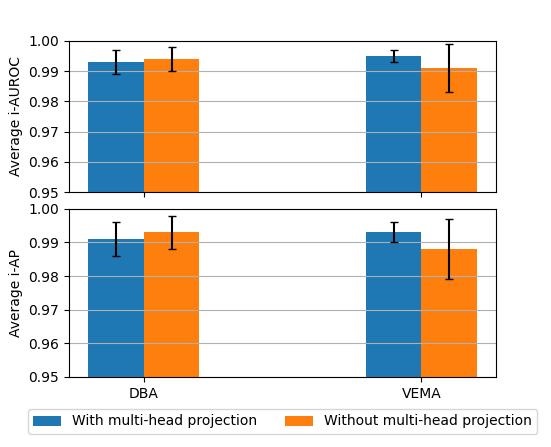}
         \caption{Key instance detection}
         \label{abl_attnmh_exp}
     \end{subfigure}
     \caption{Model performance for the signature verification task under our CAP framework (DBA and VEMA), with and without the multi-head linear projections. Left column (\ref{abl_attnmh_cls}) is classification performance; right column (\ref{abl_attnmh_exp}) shows performance of detecting key instances.}
     \label{fig:abl_attnmh}
\end{figure}

\begin{figure}[ht!]
     \centering
     \begin{subfigure}[t]{0.49\textwidth}
         \centering
         \includegraphics[width=\textwidth, left]{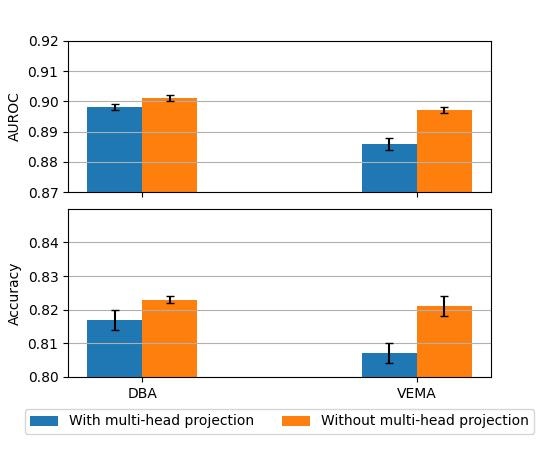}
         \caption{Classification}
         \label{abl_attnmh_cls_fever}
     \end{subfigure}
     \hfill
     \begin{subfigure}[t]{0.49\textwidth}
         \includegraphics[width=\textwidth]{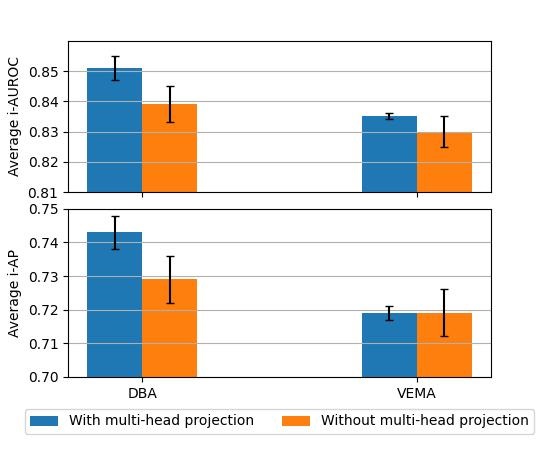}
         \caption{Key instance detection}
         \label{abl_attnmh_exp_fever}
     \end{subfigure}
     \caption{Model performance for the FEVER task under our CAP framework (DBA and VEMA), with and without the multi-head linear projections. Left column (\ref{abl_attnmh_cls_fever}) is classification performance; right column (\ref{abl_attnmh_exp_fever}) shows performance of detecting key instances.}
     \label{fig:abl_attnmh_fever}
\end{figure} 

Given the widespread popularity of  multi-head attention in the transformer literature, this section analyzes the contribution of the multi-head projection component, leaving all else intact in CAP. To attribute model performance to the multi-head mechanism, we train models with and without the linear projection to multiple heads. We also refer to attention \emph{without} multi-head projection as ``non-headed''.


Figures \ref{fig:abl_attnmh_qmnist}, \ref{fig:abl_attnmh}, \ref{fig:abl_attnmh_fever} report the out-of-sample results of the ``multi-head'' and ``non-headed'' versions of DBA and VEMA under the CAP framework, on the three tasks respectively.

Generally speaking, the differences between the multi-head and non-headed versions are small, either insignificant or marginally significant. For the CV tasks (QMNIST and signature verification), whilst multi-head VEMA obtains statistically indistinguishable improvements over the non-headed version, CAP-DBA has even slightly better performance by excluding the multi-head projection, albeit with little statistical significance. For the NLP task (FEVER), the results are inconclusive. Even though better key instance detection performance is notable for models with multi-head projection, this nevertheless does not translate to better classification---actually, we obtain significantly worse classification performance---for both CAP-VEMA and CAP-DBA (\emph{cf.} Figure \ref{abl_attnmh_cls_fever}).

\begin{table}[t]
\begin{tabular}{@{}ll c@{\hskip6pt}c@{\hskip3pt}c@{\hskip3pt}c@{\hskip3pt}c@{}}
\multicolumn{2}{c}{\multirow{2}{*}{Layer Normalization}} & \multicolumn{2}{c}{Classification} & & \multicolumn{2}{c}{Key instance detection} \\
\cline{3-4}\cline{6-7}
\multicolumn{2}{c}{} &	AUROC	&	Accuracy	&&	Avg. i-AUROC	&	Avg. i-AP \\
\toprule
\multirow{2}{*}{\emph{CAP-VEMA}} &	Post-aggregation &	0.728$\pm$0.001	&	0.672$\pm$0.003	&&	0.806$\pm$0.020	&	0.757$\pm$0.019	\\
& Pre-aggregation & \textbf{0.736$\pm$0.004} & \textbf{0.675$\pm$0.007} && \textbf{0.832$\pm$0.005} & \textbf{0.784$\pm$0.007}\\ \midrule
\multirow{2}{*}{\emph{CAP-DBA}} & Post-aggregation	& \textbf{0.738$\pm$0.005}	&	\textbf{0.684$\pm$0.006}	&&	\textbf{0.839$\pm$0.004}	& \textbf{0.790$\pm$0.004} \\
& Pre-aggregation &0.731$\pm$0.011&0.675$\pm$0.009&&0.825$\pm$0.006&0.771$\pm$0.005\\
\bottomrule 
\end{tabular}
\caption{Comparisons between pre-aggregation LayerNorm and an alternative of post-aggregation LayerNorm, on the QMNIST handwriting verification task. Higher performance metrics are bolded.}
\label{tab:ablln_qmnist}
\end{table}

\begin{table}[t]
\begin{tabular}{@{}ll c@{\hskip6pt}c@{\hskip3pt}c@{\hskip3pt}c@{\hskip3pt}c@{}}
\multicolumn{2}{c}{\multirow{2}{*}{Layer Normalization}} & \multicolumn{2}{c}{Classification} & & \multicolumn{2}{c}{Key instance detection} \\
\cline{3-4}\cline{6-7}
\multicolumn{2}{c}{} &	AUROC	&	Accuracy	&&	Avg. i-AUROC	&	Avg. i-AP \\
\toprule
\multirow{2}{*}{\emph{CAP-VEMA}} &	Post-aggregation &0.809$\pm$0.005&0.669$\pm$0.005&&0.993$\pm$0.004&0.991$\pm$0.005 \\
	& Pre-aggregation & \textbf{0.819$\pm$0.006} & \textbf{0.673$\pm$0.027} && \textbf{0.995$\pm$0.002} & \textbf{0.993$\pm$0.003} \\ \midrule
\multirow{2}{*}{\emph{CAP-DBA}} & Post-aggregation	&0.805$\pm$0.014&0.660$\pm$0.012&&0.993$\pm$0.005&0.991$\pm$0.006 \\
& Pre-aggregation &	\textbf{0.814$\pm$0.006}	& \textbf{0.674$\pm$0.018}	&&	0.993$\pm$0.004	& 0.991$\pm$0.005 \\
\bottomrule 
\end{tabular}
\caption{Comparisons between pre-aggregation LayerNorm and an alternative of post-aggregation LayerNorm, on the signature verification task. Higher performance metrics are bolded.}
\label{tab:ablln_sig}
\end{table}

\begin{table}[t!]
\begin{tabular}{@{}ll c@{\hskip6pt}c@{\hskip3pt}c@{\hskip3pt}c@{\hskip3pt}c@{}}
\multicolumn{2}{c}{\multirow{2}{*}{Layer Normalization}}  & \multicolumn{2}{c}{Classification} & & \multicolumn{2}{c}{Key instance detection} \\
\cline{3-4}\cline{6-7}
\multicolumn{2}{c}{} &AUROC&Accuracy&&Avg. i-AUROC&Avg. i-AP\\
\toprule
\multirow{2}{*}{\emph{CAP-VEMA}}&Post-aggregation &  \textbf{0.892$\pm$0.001} & \textbf{0.811$\pm$0.002}
&& \textbf{0.838$\pm$0.001} & \textbf{0.725$\pm$0.002}\\
&Pre-aggregation&0.886$\pm$0.002&0.807$\pm$0.003&&0.835$\pm$0.001&0.719$\pm$0.002\\ \midrule
\multirow{2}{*}{\emph{CAP-DBA}}&Post-aggregation&0.897$\pm$0.001&0.818$\pm$0.001&&0.838$\pm$0.006&0.728$\pm$0.007\\
&Pre-aggregation&0.898$\pm$0.001&0.817$\pm$0.003&&\textbf{0.851$\pm$0.004}&\textbf{0.743$\pm$0.005}\\
\bottomrule 
\end{tabular}
\caption{Comparisons between pre-aggregation LayerNorm and an alternative of post-aggregation LayerNorm, on the FEVER task. Higher performance metrics are bolded.}
\label{tab:ablln_fever}
\end{table}

\section{Sensitivity of our models to different forms of LayerNorm} \label{abl_ln}

One alternative to pre-aggregation LayerNorm is the post-aggregation LayerNorm adopted in the transformer model, i.e., a standard LayerNorm without multi-headedness. We replace the pre-aggregation LayerNorm with this alternative in both CAP-VEMA and CAP-DBA, and compare performance before and after replacement.

The results, shown in Tables \ref{tab:ablln_qmnist}, \ref{tab:ablln_sig}, \ref{tab:ablln_fever} for the three tasks, respectively, are mixed, across classification and key instance detection performance, and across different tasks. The performance differences between the two treatments are generally insignificant or marginally significant. The results indicate no clear winner and demonstrate that CAP-based models are not sensitive to different forms of LayerNorm when applied in the final step before output. Therefore, their performance is robust to the choice of output normalization.

\clearpage
\bibliography{references}

\begin{thebibliography}{46}
\providecommand{\natexlab}[1]{#1}
\providecommand{\url}[1]{\texttt{#1}}
\expandafter\ifx\csname urlstyle\endcsname\relax
  \providecommand{\doi}[1]{doi: #1}\else
  \providecommand{\doi}{doi: \begingroup \urlstyle{rm}\Url}\fi

\bibitem[Ba et~al.(2016)Ba, Kiros, and Hinton]{layernorm16}
Jimmy~Lei Ba, Jamie~Ryan Kiros, and Geoffrey~E. Hinton.
\newblock Layer normalization.
\newblock \emph{arXiv preprint arXiv:1607.06450}, 2016.

\bibitem[Bromley et~al.(1993)Bromley, Guyon, LeCun, S\"{a}ckinger, and Shah]{siamver93}
Jane Bromley, Isabelle Guyon, Yann LeCun, Eduard S\"{a}ckinger, and Roopak Shah.
\newblock Signature verification using a ``siamese'' time delay neural network.
\newblock In \emph{Advances in Neural Information Processing Systems}, volume~6, 1993.

\bibitem[Carbonneau et~al.(2018)Carbonneau, Cheplygina, Granger, and Gagnon]{milsurvey18}
Marc-André Carbonneau, Veronika Cheplygina, Eric Granger, and Ghyslain Gagnon.
\newblock Multiple instance learning: A survey of problem characteristics and applications.
\newblock \emph{Pattern Recognition}, 77:\penalty0 329--353, 2018.

\bibitem[Chen et~al.(2020)Chen, Kornblith, Norouzi, and Hinton]{simclr20}
Ting Chen, Simon Kornblith, Mohammad Norouzi, and Geoffrey~E. Hinton.
\newblock A simple framework for contrastive learning of visual representations.
\newblock In \emph{Proceedings of the 37th International Conference on Machine Learning, ICML 2020}, volume 119, pages 1597--1607, 2020.

\bibitem[Choe et~al.(2023)Choe, Oh, Chun, Lee, Akata, and Shim]{wsol23}
Junsuk Choe, Seong~Joon Oh, Sanghyuk Chun, Seungho Lee, Zeynep Akata, and Hyunjung Shim.
\newblock Evaluation for weakly supervised object localization: Protocol, metrics, and datasets.
\newblock \emph{IEEE Transactions on Pattern Analysis and Machine Intelligence}, 45\penalty0 (2):\penalty0 1732--1748, 2023.

\bibitem[Chopra et~al.(2005)Chopra, Hadsell, and LeCun]{siamsim05}
Sumit Chopra, Raia Hadsell, and Yann LeCun.
\newblock Learning a similarity metric discriminatively, with application to face verification.
\newblock In \emph{Proceedings of the 2005 IEEE Computer Society Conference on Computer Vision and Pattern Recognition (CVPR'05)}, volume~1, pages 539--546, 2005.

\bibitem[Conneau et~al.(2017)Conneau, Kiela, Schwenk, Barrault, and Bordes]{infersent17}
Alexis Conneau, Douwe Kiela, Holger Schwenk, Lo{\"\i}c Barrault, and Antoine Bordes.
\newblock Supervised learning of universal sentence representations from natural language inference data.
\newblock In \emph{Proceedings of the 2017 Conference on Empirical Methods in Natural Language Processing (EMNLP)}, pages 670--680, 2017.

\bibitem[Cover and Thomas(2006)]{eit06}
Thomas~M. Cover and Joy~A. Thomas.
\newblock \emph{Elements of Information Theory 2nd Edition}.
\newblock Wiley-Interscience, 2006.

\bibitem[Dauphin et~al.(2016)Dauphin, Fan, Auli, and Grangier]{glu16}
Yann Dauphin, Angela Fan, Michael Auli, and David Grangier.
\newblock Language modeling with gated convolutional networks.
\newblock In \emph{International Conference on Machine Learning}, 2016.

\bibitem[Davis and Goadrich(2006)]{prcauc06}
Jesse Davis and Mark Goadrich.
\newblock The relationship between precision-recall and roc curves.
\newblock In \emph{Proceedings of the 23rd International Conference on Machine Learning}, volume~06, pages 233--240, 2006.

\bibitem[Dietterich et~al.(1997)Dietterich, Lathrop, and Lozano-Pérez]{mil97}
Thomas~G. Dietterich, Richard~H. Lathrop, and Tomás Lozano-Pérez.
\newblock Solving the multiple instance problem with axis-parallel rectangles.
\newblock \emph{Artificial Intelligence}, 89\penalty0 (1):\penalty0 31--71, 1997.

\bibitem[Dosovitskiy et~al.(2021)Dosovitskiy, Beyer, Kolesnikov, Weissenborn, Zhai, Unterthiner, Dehghani, Minderer, Heigold, Gelly, Uszkoreit, and Houlsby]{vit21}
Alexey Dosovitskiy, Lucas Beyer, Alexander Kolesnikov, Dirk Weissenborn, Xiaohua Zhai, Thomas Unterthiner, Mostafa Dehghani, Matthias Minderer, Georg Heigold, Sylvain Gelly, Jakob Uszkoreit, and Neil Houlsby.
\newblock An image is worth 16x16 words: Transformers for image recognition at scale.
\newblock In \emph{International Conference on Learning Representations}, 2021.

\bibitem[Early et~al.(2022)Early, Evers, and Ramchurn]{milinterpret22}
Joseph Early, Christine Evers, and Sarvapali Ramchurn.
\newblock Model agnostic interpretability for multiple instance learning.
\newblock In \emph{ICLR}, 2022.

\bibitem[Ghosh et~al.(2023)Ghosh, Lyu, Zhang, and Wang]{resnet18mnist23}
Avrajit Ghosh, He~Lyu, Xitong Zhang, and Rongrong Wang.
\newblock Implicit regularization in heavy-ball momentum accelerated stochastic gradient descent.
\newblock In \emph{The Eleventh International Conference on Learning Representations (ICLR)}, 2023.

\bibitem[He et~al.(2016)He, Zhang, Ren, and Sun]{resnet15}
Kaiming He, Xiangyu Zhang, Shaoqing Ren, and Jian Sun.
\newblock Deep residual learning for image recognition.
\newblock In \emph{2016 IEEE Conference on Computer Vision and Pattern Recognition (CVPR)}, pages 770--778, 2016.

\bibitem[Hsieh et~al.(2019)Hsieh, Lo, Chen, and Liu]{sce19}
Ting{-}I Hsieh, Yi{-}Chen Lo, Hwann{-}Tzong Chen, and Tyng{-}Luh Liu.
\newblock One-shot object detection with co-attention and co-excitation.
\newblock In \emph{Advances in Neural Information Processing Systems}, 2019.

\bibitem[Hu et~al.(2018)Hu, Shen, and Sun]{se18}
Jie Hu, Li~Shen, and Gang Sun.
\newblock Squeeze-and-excitation networks.
\newblock In \emph{Computer Vision and Pattern Recognition}, pages 7132--7141, 2018.

\bibitem[Ilse et~al.(2018)Ilse, Tomczak, and Welling]{gabmil18}
Maximilian Ilse, Jakub~M. Tomczak, and M.~Welling.
\newblock Attention-based deep multiple instance learning.
\newblock In \emph{Proceedings of the 35th International Conference on Machine Learning}, pages 2127--2136, 2018.

\bibitem[Jia et~al.(2021)Jia, Frank, Pfahringer, Bifet, and Lim]{explain21}
Yunzhe Jia, Eibe Frank, Bernhard Pfahringer, Albert Bifet, and Nick Jin~Sean Lim.
\newblock Studying and exploiting the relationship between model accuracy and explanation quality.
\newblock In \emph{ECML-PKDD}, 2021.

\bibitem[Kanavati and Tsuneki(2021)]{bn21}
Fahdi Kanavati and Masayuki Tsuneki.
\newblock Partial transfusion: on the expressive influence of trainable batch norm parameters for transfer learning.
\newblock In \emph{Proceedings of Machine Learning Research}, volume 143, pages 338--353, 2021.

\bibitem[Koch et~al.(2015)Koch, Zemel, and Salakhutdinov]{siamese15}
Gregory Koch, Richard Zemel, and Ruslan Salakhutdinov.
\newblock Siamese neural networks for one-shot image recognition.
\newblock In \emph{Proceedings of the 32nd International Conference on Machine Learning}, 2015.

\bibitem[Lee et~al.(2019)Lee, Lee, Kim, Kosiorek, Choi, and Teh]{pma19}
Juho Lee, Yoonho Lee, Jungtaek Kim, Adam~R. Kosiorek, Seungjin Choi, and Yee~Whye Teh.
\newblock Set transformer: A framework for attention-based permutation-invariant neural networks.
\newblock In \emph{Proceedings of the 36th International Conference on Machine Learning}, pages 3744--3753, 2019.

\bibitem[Li and Vasconcelos(2015)]{milsoftbag15}
Weixin Li and Nuno Vasconcelos.
\newblock Multiple instance learning for soft bags via top instances.
\newblock In \emph{2015 IEEE Conference on Computer Vision and Pattern Recognition (CVPR)}, pages 4277--4285, 2015.

\bibitem[Liu et~al.(2019)Liu, Bi, Ma, and Wang]{micnn19}
Xiaokai Liu, Sheng Bi, Xiaorui Ma, and Jie Wang.
\newblock Multi-instance convolutional neural network for multi-shot person re-identification.
\newblock \emph{Neurocomputing}, 337:\penalty0 303--314, 2019.

\bibitem[Liwicki et~al.(2011)Liwicki, Blumenstein, van~den Heuvel, Berger, Stoel, Found, Chen, and Malik]{sigcomp11}
Marcus Liwicki, Michael Blumenstein, Elisa van~den Heuvel, Charles~E.H. Berger, Reinoud~D. Stoel, Bryan Found, Xiaohong Chen, and Muhammad~Imran Malik.
\newblock Sigcomp11: Signature verification competition for on- and offline skilled forgeries.
\newblock In \emph{Proc. 11th Int. Conference on Document analysis and Recognition}, 2011.

\bibitem[Mukhoti et~al.(2021)Mukhoti, Kirsch, van Amersfoort, Torr, and Gal]{dirtymnist21}
Jishnu Mukhoti, Andreas Kirsch, Joost van Amersfoort, Philip H.~S. Torr, and Yarin Gal.
\newblock Deterministic neural networks with appropriate inductive biases capture epistemic and aleatoric uncertainty.
\newblock \emph{arXiv preprint arXiv:2102.11582}, 2021.

\bibitem[Park et~al.(2023)Park, Shin, Hwang, and Choi]{ambmnist23}
Jeongeun Park, Seungyoun Shin, Sangheum Hwang, and Sungjoon Choi.
\newblock Elucidating robust learning with uncertainty-aware corruption pattern estimation.
\newblock \emph{Pattern Recognition}, 138:\penalty0 109387, 2023.

\bibitem[Pedregosa et~al.(2011)Pedregosa, Varoquaux, Gramfort, Michel, Thirion, Grisel, Blondel, Prettenhofer, Weiss, Dubourg, Vanderplas, Passos, Cournapeau, Brucher, Perrot, and Duchesnay]{sklearn}
F.~Pedregosa, G.~Varoquaux, A.~Gramfort, V.~Michel, B.~Thirion, O.~Grisel, M.~Blondel, P.~Prettenhofer, R.~Weiss, V.~Dubourg, J.~Vanderplas, A.~Passos, D.~Cournapeau, M.~Brucher, M.~Perrot, and E.~Duchesnay.
\newblock Scikit-learn: Machine learning in {P}ython.
\newblock \emph{Journal of Machine Learning Research}, 12:\penalty0 2825--2830, 2011.

\bibitem[Reimers and Gurevych(2019)]{sbert19}
Nils Reimers and Iryna Gurevych.
\newblock Sentence-bert: Sentence embeddings using siamese bert-networks.
\newblock In \emph{EMNLP}, 2019.

\bibitem[Sathe and Park(2021)]{fever_mil21}
Aalok Sathe and Joonsuk Park.
\newblock Automatic fact-checking with document-level annotations using {BERT} and multiple instance learning.
\newblock In \emph{Proceedings of the Fourth Workshop on Fact Extraction and VERification (FEVER)}, pages 101--107. Association for Computational Linguistics, 2021.

\bibitem[Shao et~al.(2021)Shao, Bian, Chen, Wang, Zhang, Ji, and Zhang]{transmil21}
Zhucheng Shao, Hao Bian, Yang Chen, Yifeng Wang, Jian Zhang, Xiangyang Ji, and Yongbing Zhang.
\newblock {TransMIL}: Transformer based correlated multiple instance learning for whole slide image classication.
\newblock In \emph{Advances in Neural Information Processing Systems}, 2021.

\bibitem[Tan and Le(2021)]{effnet21}
Mingxing Tan and Quoc~V. Le.
\newblock {EfficientNetV2}: Smaller models and faster training.
\newblock In \emph{International Conference on Machine Learning}, 2021.

\bibitem[Tarek et~al.(2022)Tarek, Hamouda, and Abohamama]{mibiomgan22}
M.~Tarek, E.~Hamouda, and A.S. Abohamama.
\newblock Multi-instance cancellable biometrics schemes based on generative adversarial network.
\newblock \emph{Applied Intelligence}, 52\penalty0 (1):\penalty0 501--513, 2022.

\bibitem[Thorne et~al.(2018)Thorne, Vlachos, Christodoulopoulos, and Mittal]{fever18}
James Thorne, Andreas Vlachos, Christos Christodoulopoulos, and Arpit Mittal.
\newblock {FEVER}: a large-scale dataset for fact extraction and {VERification}.
\newblock In \emph{NAACL-HLT}, 2018.

\bibitem[Vaswani et~al.(2017)Vaswani, Shazeer, Parmar, Uszkoreit, Jones, Gomez, Kaiser, and Polosukhin]{transformer17}
Ashish Vaswani, Noam Shazeer, Niki Parmar, Jakob Uszkoreit, Llion Jones, Aidan~N. Gomez, Lukasz Kaiser, and Illia Polosukhin.
\newblock Attention is all you need.
\newblock In \emph{Advances in Neural Information Processing Systems}, pages 5998--6008, 2017.

\bibitem[Wang et~al.(2020)Wang, M., and Tuytelaars]{mil-lstm20}
Kaili Wang, Jos{\'{e}}~Oramas M., and Tinne Tuytelaars.
\newblock In defense of {LSTM}s for addressing multiple instance learning problems.
\newblock In \emph{15th ACCV}, pages 444--460, 2020.

\bibitem[Wang et~al.(2018)Wang, Yan, Tang, Bai, and Liu]{minet18}
Xinggang Wang, Yongluan Yan, Peng Tang, Xiang Bai, and Wenyu Liu.
\newblock Revisiting multiple instance neural networks.
\newblock \emph{Pattern Recognition}, 74:\penalty0 15--24, 2018.

\bibitem[Xu and Frank(2004)]{miboost04}
Xin Xu and Eibe Frank.
\newblock Logistic regression and boosting for labeled bags of instances.
\newblock In \emph{Advances in Knowledge Discovery and Data Mining}, pages 272--281, 2004.

\bibitem[Yadav and Bottou(2019)]{qmnist19}
Chhavi Yadav and L\'{e}on Bottou.
\newblock Cold case: The lost mnist digits.
\newblock In \emph{Advances in Neural Information Processing Systems 32}, 2019.

\bibitem[Yan et~al.(2018)Yan, Wang, Guo, Fang, Liu, and Huang]{mildp18}
Yongluan Yan, Xinggang Wang, Xiaojie Guo, Jiemin Fang, Wenyu Liu, and Junzhou Huang.
\newblock Deep multi-instance learning with dynamic pooling.
\newblock In \emph{Proceedings of The 10th Asian Conference on Machine Learning}, volume~95, pages 662--677, 2018.

\bibitem[Yu et~al.(2020)Yu, Chen, Du, Li, Rashwan, Hou, Jin, Yang, Liu, Kim, and Li]{tf_model20}
Hongkun Yu, Chen Chen, Xianzhi Du, Yeqing Li, Abdullah Rashwan, Le~Hou, Pengchong Jin, Fan Yang, Frederick Liu, Jaeyoun Kim, and Jing Li.
\newblock {TensorFlow Model Garden}.
\newblock \url{https://github.com/tensorflow/models}, 2020.

\bibitem[Zagoruyko and Komodakis(2015)]{siamcnn15}
Sergey Zagoruyko and Nikos Komodakis.
\newblock Learning to compare image patches via convolutional neural networks.
\newblock In \emph{Proceedings of the IEEE Conference on Computer Vision and Pattern Recognition (CVPR)}, June 2015.

\bibitem[Zbontar et~al.(2021)Zbontar, Jing, Misra, LeCun, and Deny]{barlow21}
Jure Zbontar, Li~Jing, Ishan Misra, Yann LeCun, and St{\'{e}}phane Deny.
\newblock Barlow twins: Self-supervised learning via redundancy reduction.
\newblock In \emph{Proceedings of the 38th International Conference on Machine Learning, {ICML} 2021}, volume 139, pages 12310--12320, 2021.

\bibitem[Zhang and Goldman(2001)]{emdd02}
Qi~Zhang and Sally Goldman.
\newblock {EM-DD}: An improved multiple-instance learning technique.
\newblock In \emph{Advances in Neural Information Processing Systems}, volume~14, 2001.

\bibitem[Zhou(2017)]{milreview17}
Zhi-Hua Zhou.
\newblock {A brief introduction to weakly supervised learning}.
\newblock \emph{National Science Review}, 5\penalty0 (1):\penalty0 44--53, 2017.

\bibitem[Zhou et~al.(2009)Zhou, Sun, and Li]{milcorr09}
Zhi{-}Hua Zhou, Yu{-}Yin Sun, and Yu{-}Feng Li.
\newblock Multi-instance learning by treating instances as non-i.i.d. samples.
\newblock In \emph{International Conference on Machine Learning}, pages 1249--1256, 2009.

\end{thebibliography}

\end{document}